\documentclass[11pt]{article}
\usepackage[preprint]{acl}
\usepackage{times}
\usepackage{latexsym}
\usepackage{type1cm} %
\usepackage{etoolbox}
\usepackage[T1]{fontenc}
\usepackage[utf8]{inputenc}
\usepackage{microtype}
\usepackage{inconsolata}
\usepackage{amsmath,amssymb}
\usepackage{graphicx}
\usepackage{tikz}
\usepackage{booktabs}
\usepackage{colortbl}
\usepackage{tabularx}
\usepackage{multirow}
\usepackage{arydshln}
\usepackage{xurl}
\usepackage{placeins}
\definecolor{sourceAPIColor}{HTML}{377BC2}
\definecolor{sourcePublicColor}{HTML}{20958B}
\definecolor{sourceInhouseColor}{HTML}{D4832F}
\DeclareRobustCommand{\sourceAPI}{%
  \tikz[baseline={([yshift=-0.75ex]current bounding box.center)},x=9pt,y=9pt,draw=sourceAPIColor,line width=0.55pt,line cap=round,line join=round]{%
    \path[use as bounding box] (0,0) rectangle (1,0.85);
    \draw[fill=sourceAPIColor!15] (0.24,0.15) .. controls (-0.03,0.15) and (-0.03,0.52) .. (0.24,0.54)
      .. controls (0.24,0.87) and (0.70,0.89) .. (0.77,0.57)
      .. controls (1.07,0.59) and (1.11,0.15) .. (0.80,0.15) -- cycle;
  }%
}
\DeclareRobustCommand{\sourcePublic}{%
  \tikz[baseline={([yshift=-0.75ex]current bounding box.center)},x=9pt,y=9pt,draw=sourcePublicColor,line width=0.65pt,line cap=round,line join=round]{%
    \path[use as bounding box] (0,0) rectangle (1,0.85);
    \draw (0.5,0.8) -- (0.5,0.27) (0.28,0.48) -- (0.5,0.26) -- (0.72,0.48);
    \draw (0.12,0.29) -- (0.12,0.08) -- (0.88,0.08) -- (0.88,0.29);
  }%
}
\DeclareRobustCommand{\sourceInhouse}{%
  \tikz[baseline={([yshift=-0.75ex]current bounding box.center)},x=9pt,y=9pt,draw=sourceInhouseColor,line width=0.5pt,line join=round]{%
    \path[use as bounding box] (0,0) rectangle (1,0.85);
    \begin{scope}[shift={(0.5,0.43)}]
      \draw[fill=sourceInhouseColor!18] (0:0.31) \foreach \i in {1,...,32} {
        -- ({\i*11.25}:{ifthenelse(mod(\i,4)<2,0.31,0.43)})} -- cycle;
      \draw[fill=white] (0,0) circle (0.14);
    \end{scope}
  }%
}
\ifdefined\pdfrunninglinkoff
  \ExplSyntaxOn
  \cs_if_exist:NF \pdfannot_link_off:
    { \cs_new_eq:NN \pdfannot_link_off: \pdfrunninglinkoff }
  \cs_if_exist:NF \pdfannot_link_on:
    { \cs_new_eq:NN \pdfannot_link_on: \pdfrunninglinkon }
  \ExplSyntaxOff
  \makeatletter
  \patchcmd{\@comflelt}{\box #1}{\pdfrunninglinkoff\box #1\pdfrunninglinkon}
    {}{\PackageError{crossenv}{Cannot guard single-column float links}{}}
  \patchcmd{\@comdblflelt}{\box #1}{\pdfrunninglinkoff\box #1\pdfrunninglinkon}
    {}{\PackageError{crossenv}{Cannot guard double-column float links}{}}
  \makeatother
  \newcommand{\CrossEnvGuardShipoutLinks}{%
    \setbox\ShipoutBox=\vbox{%
      \pdfrunninglinkoff
      \box\ShipoutBox
      \pdfrunninglinkoff
    }%
  }
  \AddToHook{shipout/before}[crossenv/link-guards]{\CrossEnvGuardShipoutLinks}
  \AddToHook{shipout}[crossenv/link-guards]{\CrossEnvGuardShipoutLinks}
\fi

\hypersetup{
  pdftitle={CrossEnv: Synthesizing Verifiable Cross-Environment Tasks for Long-Horizon Agent Training},
  pdfauthor={Zhengfeng Yu, Tian Lan, Zhijing Wu, Enbo Zhao, Xuannan Liu, Feihu Jiang, Qianghuai Jia, Longyue Wang, Zhao Xu, Weihua Luo, Xian-Ling Mao}
}
\title{\texorpdfstring{\raisebox{-0.25em}{\includegraphics[height=1.2em]{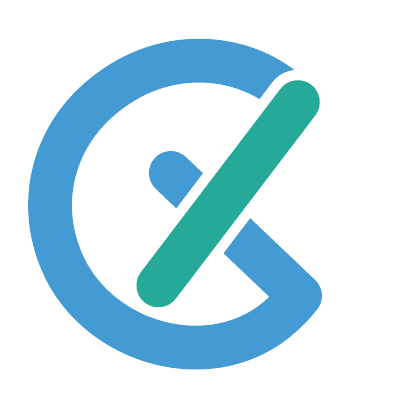}}\hspace{0.3em}}{}CrossEnv: Synthesizing Verifiable Cross-Environment Tasks for Long-Horizon Agent Training}
\author{
  \textbf{Zhengfeng Yu$^{1,2}$} \quad
  \textbf{Tian Lan$^{2}$} \quad
  \textbf{Zhijing Wu$^{1}$}\thanks{Corresponding author.} \quad
  \textbf{Enbo Zhao$^{2}$} \\
  \textbf{Xuannan Liu$^{3}$} \quad
  \textbf{Feihu Jiang$^{2}$} \quad
  \textbf{Qianghuai Jia$^{2}$} \quad
  \textbf{Longyue Wang$^{2}$} \\
  \textbf{Zhao Xu$^{2}$} \quad
  \textbf{Weihua Luo$^{2}$} \quad
  \textbf{Xian-Ling Mao$^{1}$} \\
  \normalfont $^{1}$ Beijing Institute of Technology \\
  \normalfont $^{2}$ Alibaba Group \\
  \normalfont $^{3}$ Beijing University of Posts and Telecommunications
}
\begin{document}
\begin{NoHyper}
\maketitle
\end{NoHyper}

\begin{abstract}

Executable synthetic environments provide scalable training tasks and verifiable feedback for long-horizon agents. Existing environment synthesis methods primarily generate tasks within individual environments. However, these tasks do not adequately capture real-world interactions across services, where actions in one service depend on information obtained from another. Synthesizing such tasks is challenging because data generated across environments must remain consistent to support executable and verifiable interactions. To address this challenge, we introduce \textbf{CrossEnv}, a framework that uses an entity--tool graph to identify connections between environments and aligns task-specific data to synthesize executable, verifiable cross-environment tasks. We further propose \textbf{CrossEnv-Curriculum}, which progresses from single-environment tool use to coordination across more environments and business domains, before introducing user interaction. Across four external benchmarks, combining cross-environment training data with this curriculum achieves competitive performance with fewer training examples. Further analyses support the benefits of cross-environment data and curriculum training, and provide evidence of transfer to new environment combinations, external tasks less similar to the training data, and longer workflows.

\end{abstract}
\section{Introduction}

Large language models (LLMs) are increasingly used as agents that invoke tools~\citep{schick2023toolformerlanguagemodelsteach} and adapt their actions based on environmental feedback~\citep{yao2023reactsynergizingreasoningacting,liu2025agentbenchevaluatingllmsagents,shinn2023reflexion}. Training these agents requires diverse executable environments and verifiable tasks that provide reliable feedback~\citep{xi2025agentgymrltrainingllmagents}. However, restricted access and high interaction costs limit training in real-world applications~\citep{chen2025scalingagentlearningexperience}. Programmatic environment synthesis offers a scalable alternative by generating executable tools, stateful environments, and verifiable training tasks~\citep{song2026envscalerscalingtoolinteractiveenvironments,tu2026scaleenvscalingenvironmentsynthesis}.

\begin{figure}[t]
\centering
\includegraphics[width=\columnwidth]{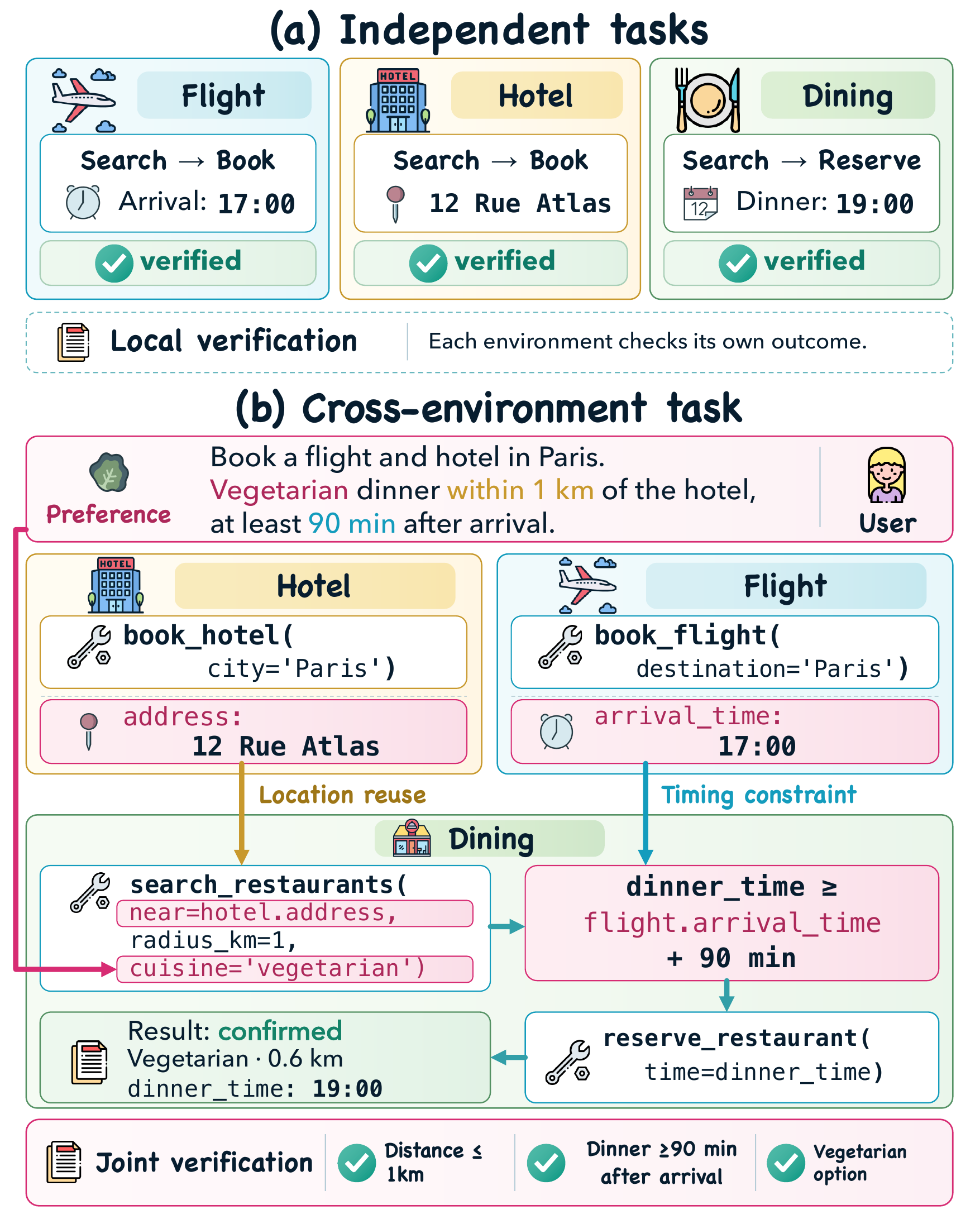}
\caption{Independent tasks and a cross-environment task. (a) Flight, hotel, and dining tasks are completed and verified independently. (b) A travel-planning task uses flight arrival time, hotel location, and personal preferences to guide restaurant reservation, with joint verification of cross-environment requirements.}
\label{fig:motivation}
\end{figure}

Real-world long-horizon workflows often span multiple services, where actions in one service depend on information obtained from another~\citep{trivedi2024appworldcontrollableworldapps,li2026tooldecathlonbenchmarkinglanguage,xie2024osworld}. Figure~\ref{fig:motivation}(b) illustrates a travel-planning example: after booking flights and accommodation, an agent selects and reserves a restaurant based on the flight arrival time, hotel location, and personal preferences. Training agents for such workflows requires tasks that capture these information dependencies and the coordinated actions needed to achieve a shared goal.

However, existing programmatic environment synthesis methods primarily generate single-environment tasks, with limited support for combining independently constructed environments into joint tasks~\citep{song2026envscalerscalingtoolinteractiveenvironments,wang2026agentworldmodelinfinity}. Cross-application benchmarks capture some cross-environment dependencies, but task and evaluator design is often manual or semi-automated~\citep{trivedi2024appworldcontrollableworldapps,li2026tooldecathlonbenchmarkinglanguage,xu-etal-2025-crab,zhou2024webarenarealisticwebenvironment}, limiting their scalability for training. This leaves a gap in automatically synthesizing cross-environment training tasks at scale.

A key difficulty is maintaining \textbf{data consistency across environments}. Data constructed independently for different environments may be valid in isolation but inconsistent with one another. For example, the same hotel may be recorded with different addresses in the booking and navigation environments. Synthesizing cross-environment tasks therefore requires aligning relevant data across environments to support executable, verifiable tasks.

To address this challenge, we introduce \textbf{CrossEnv}, a framework for automatically synthesizing executable, verifiable cross-environment tasks. CrossEnv uses an entity--tool graph to connect environments through shared entity types and guide the construction of task workflows. It then aligns task-specific data across environments to support execution and generates verification rules from task objectives and reference execution.

CrossEnv generates tasks with varying coordination complexity by controlling the number of participating environments and their business-domain scope. We introduce \textbf{CrossEnv-Curriculum} to establish basic tool-use skills before progressively training agents to coordinate operations across environments. Training begins with supervised fine-tuning on single-environment reference trajectories, followed by stagewise reinforcement learning that expands to multiple environments within a business domain, then across domains, and finally introduces user interaction.

\begin{table}[!t]
\centering
\newcommand{\envbrush}[2]{%
  \textcolor[HTML]{#1}{\tikz[x=1pt,y=1pt,baseline=-0.7pt]{%
    \path[use as bounding box] (0,-0.2) rectangle (6,6.4);
    #2%
  }}%
}
\DeclareRobustCommand{\envyes}{\envbrush{45982C}{%
  \fill (0.35,2.45)
    .. controls (0.68,2.98) and (1.02,3.02) .. (1.30,2.55)
    .. controls (1.52,2.16) and (1.69,1.80) .. (1.88,1.49)
    .. controls (2.81,3.18) and (4.18,5.10) .. (5.43,6.04)
    .. controls (5.65,6.21) and (5.84,6.21) .. (5.98,6.11)
    .. controls (4.66,4.54) and (3.18,2.37) .. (2.30,0.49)
    .. controls (2.10,0.10) and (1.39,0.02) .. (1.18,0.42)
    .. controls (0.88,1.02) and (0.61,1.79) .. (0.35,2.45) -- cycle;
}}
\DeclareRobustCommand{\envpart}{\envbrush{F2AA3D}{%
  \fill[even odd rule] (0.23,0.38)
    .. controls (0.61,1.43) and (1.93,4.61) .. (2.51,5.92)
    .. controls (2.64,6.23) and (2.94,6.18) .. (3.10,5.85)
    .. controls (3.80,4.32) and (4.86,1.77) .. (5.57,0.49)
    .. controls (5.72,0.20) and (5.52,0.04) .. (5.22,0.10)
    .. controls (3.87,0.26) and (1.81,0.04) .. (0.56,0.09)
    .. controls (0.31,0.09) and (0.14,0.15) .. (0.23,0.38) -- cycle
    (1.36,0.94)
    .. controls (2.22,0.90) and (3.56,1.06) .. (4.46,0.91)
    .. controls (3.99,1.98) and (3.23,3.84) .. (2.84,4.69)
    .. controls (2.37,3.55) and (1.70,1.81) .. (1.36,0.94) -- cycle;
}}
\DeclareRobustCommand{\envno}{\envbrush{A32217}{%
  \fill (1.03,5.65)
    .. controls (0.72,5.94) and (0.50,5.60) .. (0.70,5.13)
    .. controls (1.59,3.54) and (2.78,1.70) .. (3.95,0.51)
    .. controls (4.28,0.15) and (4.73,0.43) .. (4.58,0.84)
    .. controls (3.54,2.37) and (2.45,4.32) .. (1.43,5.95)
    .. controls (1.28,6.17) and (1.15,5.90) .. (1.03,5.65) -- cycle;
  \fill (4.34,5.85)
    .. controls (3.08,4.49) and (1.89,2.35) .. (0.93,0.48)
    .. controls (0.72,0.03) and (1.14,-0.08) .. (1.43,0.24)
    .. controls (2.73,1.82) and (3.99,3.78) .. (5.13,5.50)
    .. controls (5.36,5.87) and (5.02,6.04) .. (4.81,5.84)
    .. controls (4.67,6.09) and (4.52,6.02) .. (4.34,5.85) -- cycle;
}}
\begingroup
\fontsize{8}{9.5}\selectfont
\setlength{\tabcolsep}{0.7pt}
\renewcommand{\arraystretch}{1.1}
\newcommand{\envhead}[1]{{\fontsize{7}{8}\selectfont\bfseries\shortstack{#1}}}
\begin{tabularx}{\linewidth}{@{}>{\fontsize{7.5}{9}\selectfont\raggedright\arraybackslash}p{0.31\linewidth}*{6}{>{\centering\arraybackslash}X}@{}}
\toprule
\multirow{2}{*}[-3pt]{\textbf{Method}} & \multicolumn{3}{c}{\textbf{\fontsize{7.5}{9}\selectfont General synthesis}} & \multicolumn{3}{c}{\textbf{\fontsize{7.5}{9}\selectfont Cross-environment}} \\
\cmidrule(lr){2-4}\cmidrule(l){5-7}
 & \envhead{Exec.} & \envhead{Verif.} & \envhead{Ctrl.} & \envhead{Cross} & \envhead{Consist.} & \envhead{Ext.} \\
\midrule
\multicolumn{7}{c}{\textit{Real-world environments}} \\
\midrule
ToolLLM \citeyearpar{qin2023toolllmfacilitatinglargelanguage} & \envyes & \envno & \envyes & \envyes & \envno & \envyes \\
Toucan \citeyearpar{xu2025toucansynthesizing15mtoolagentic} & \envyes & \envno & \envyes & \envyes & \envno & \envyes \\
AgentTrek \citeyearpar{xu2025agenttrekagenttrajectorysynthesis} & \envyes & \envno & \envno & \envno & \envno & \envno \\
\midrule
\multicolumn{7}{c}{\textit{LLM-simulated environments}} \\
\midrule
Simia \citeyearpar{li2025simulatingenvironmentsreasoningmodels} & \envno & \envno & \envpart & \envyes & \envno & \envno \\
DreamGym \citeyearpar{chen2025scalingagentlearningexperience} & \envno & \envno & \envpart & \envno & \envno & \envno \\
EnvACE \citeyearpar{xu2026envaceinternalizingenvironmentdynamics} & \envno & \envno & \envno & \envno & \envno & \envno \\
\midrule
\multicolumn{7}{c}{\textit{Programmatic environments}} \\
\midrule
EnvScaler \citeyearpar{song2026envscalerscalingtoolinteractiveenvironments} & \envyes & \envyes & \envno & \envno & \envno & \envno \\
AWM \citeyearpar{wang2026agentworldmodelinfinity} & \envyes & \envyes & \envno & \envpart & \envno & \envno \\
ScaleEnv \citeyearpar{tu2026scaleenvscalingenvironmentsynthesis} & \envyes & \envyes & \envyes & \envno & \envno & \envno \\
\mbox{Agent-World} \citeyearpar{dong2026agentworldscalingrealworldenvironment} & \envyes & \envyes & \envyes & \envno & \envno & \envno \\
Skill2Env \citeyearpar{xu2026reinforcingcollectiveskills} & \envyes & \envyes & \envyes & \envno & \envno & \envno \\
\midrule
\rowcolor{black!4}
\textbf{CrossEnv} & \envyes & \envyes & \envyes & \envyes & \envyes & \envyes \\
\bottomrule
\end{tabularx}
\endgroup
\caption{Environment and task synthesis capabilities. Exec.: real-service or code execution; Verif.: programmatic outcome checks; Ctrl.: adjustable complexity; Cross: multi-environment tasks; Consist.: cross-environment data alignment; Ext.: new-environment integration into joint tasks. \envyes: supported; \envpart: limited; \envno: not explicitly provided.}
\label{tab:environment_comparison}
\end{table}

Experiments on four external benchmarks show that combining cross-environment training data with this curriculum achieves competitive performance with fewer training examples than comparable environment-trained baselines. Further analyses support the benefits of cross-environment data and curriculum training, and provide evidence of transfer to new environment combinations, external environments, and longer workflows.

Our main contributions are as follows:

\begin{enumerate}
\item We introduce \textbf{CrossEnv}, which combines entity--tool graph-guided workflows with cross-environment data alignment to synthesize executable, verifiable training tasks.

\item We propose \textbf{CrossEnv-Curriculum}, which organizes training from single-environment tool use to increasingly complex coordination across environments and business domains, and finally to tasks requiring user interaction.

\item We analyze stagewise capability development and retention, comparing curriculum versus mixed training and single- versus cross-environment tasks to assess the roles of training strategy and task composition.
\end{enumerate}

\begin{figure*}[!t]
\centering
\includegraphics[width=\linewidth]{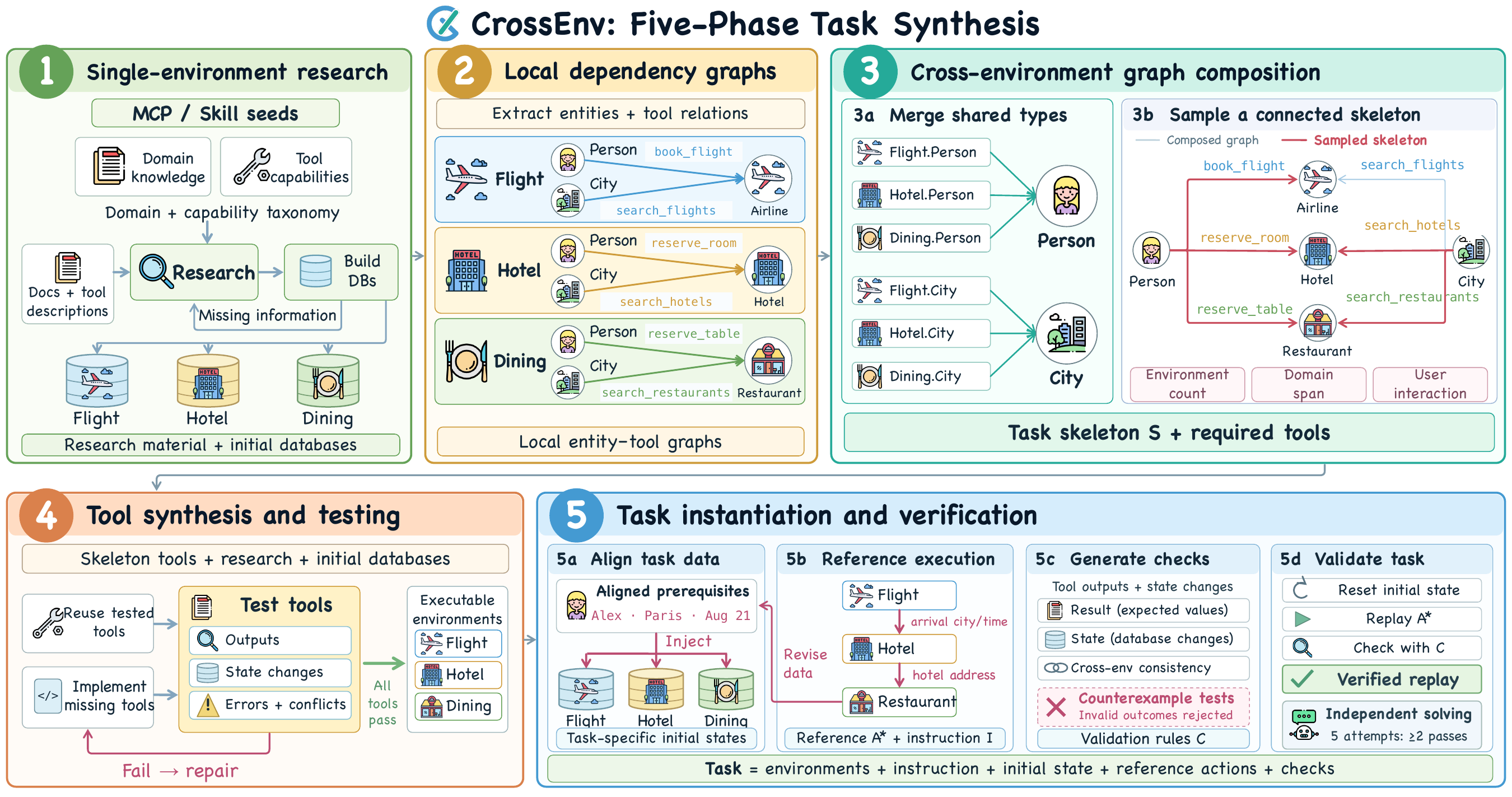}
\caption{Overview of the CrossEnv task synthesis framework. Entity--tool graph composition forms task skeletons, while tool synthesis, data alignment, and reference execution produce executable tasks with verifiable outcomes. Tasks are validated through reference replay and independent agent attempts.}
\label{fig:pipeline}
\end{figure*}

\section{Related Work}

\subsection{Environment Synthesis}

Table~\ref{tab:environment_comparison} compares approaches based on real-world access~\citep{xu2025agenttrekagenttrajectorysynthesis}, LLM simulation~\citep{chen2025scalingagentlearningexperience,xu2026envaceinternalizingenvironmentdynamics}, and programmatic construction. For general synthesis, real services and executable programs support action execution, while programmatic approaches provide task-outcome checks~\citep{song2026envscalerscalingtoolinteractiveenvironments,yang2025swesmithscalingdatasoftware,xu2026envfactoryscalingtooluseagents}. ScaleEnv~\citep{tu2026scaleenvscalingenvironmentsynthesis}, Agent-World~\citep{dong2026agentworldscalingrealworldenvironment}, and Skill2Env~\citep{xu2026reinforcingcollectiveskills} explicitly control task complexity. Graph-guided synthesis supports long-horizon tool tasks in ToolVerse~\citep{zhou2026toolverseunlockingmassiveenvironments}, terminal skill composition in SkillSynth~\citep{fan2026scalableterminaltasksynthesis}, and controllable subtask complexity in OmniBench~\citep{pmlr-v267-bu25b}. Across environments, ToolLLM~\citep{qin2023toolllmfacilitatinglargelanguage} and Toucan~\citep{xu2025toucansynthesizing15mtoolagentic} support multi-service tasks and new-service integration; Simia~\citep{li2025simulatingenvironmentsreasoningmodels} simulates multi-environment workflows, while AWM~\citep{wang2026agentworldmodelinfinity} provides limited support within bundled service suites. ARE~\citep{froger2025arescalingagentenvironments} addresses cross-app data coherence through dependency-aware generation. CrossEnv composes independently constructed environments through shared entity types and aligns task-specific records for jointly verifiable workflows.

\subsection{Curriculum Learning}

Curriculum learning organizes training examples by difficulty or learning progress to support progressive skill acquisition~\citep{Bengio_2009,NIPS2010_e57c6b95,pmlr-v70-graves17a,pmlr-v78-florensa17a,pmlr-v80-florensa18a}. For LLM agents, AgentGym-RL~\citep{xi2025agentgymrltrainingllmagents} gradually increases interaction budgets, while WebRL~\citep{qi2025webrltrainingllmweb} generates tasks from failures and filters them by difficulty. AgentGen~\citep{hu2025agentgenenhancingplanningabilities} constructs planning tasks at varying difficulty levels through bidirectional evolution. More recently, GenEnv~\citep{guo2025genenvdifficultyalignedcoevolutionllm} and Tool-R0~\citep{acikgoz2026toolr0selfevolvingllmagents} use co-evolution to generate tasks adapted to the agent's current capabilities. CrossEnv-Curriculum organizes training around the coordination structure of cross-environment workflows, providing an explicit progression from foundational execution skills to joint task completion.

\section{CrossEnv}
\label{sec:crossenv-method}

CrossEnv automatically synthesizes executable, verifiable cross-environment tasks. Figure~\ref{fig:pipeline} presents an overview of its five-phase pipeline.

A cross-environment task involves tool use in at least two environments. Task completion requires using information from one environment to guide actions in another or meeting constraints on their combined results. Each task is represented as
\begin{equation}
\label{eq:crossenv-task}
\tau=(\mathcal E_S,I,\mathbf s_{0,\tau},A^*,C_\tau),
\end{equation}
where $\mathcal E_S$ is the set of environments used in the task, $I$ is the instruction, $\mathbf s_{0,\tau}$ is their joint initial state, $A^*$ is a reference execution sequence, and $C_\tau$ contains executable verification rules. Task completion requires all rules to pass, evaluated from tool outputs and final environment states.

\subsection{Single-Environment Research}

\noindent\textbf{Environment research.} We collect environment seeds from MCP services and public skill documents, retaining 801 MCP seeds and 682 skill seeds after filtering. A research agent retrieves documentation and data, while a data construction agent builds initial databases and requests further retrieval when needed. These materials support entity extraction and tool implementation.

\noindent\textbf{Environment taxonomy.} We classify the seeds into 10 business domains and 84 capability categories, drawing on the Agent-World taxonomy~\citep{dong2026agentworldscalingrealworldenvironment} and seed functionality. This taxonomy supports within-domain and cross-domain task composition; the complete classification is provided in Appendix~\ref{app:environment-taxonomy}.

\subsection{Local Dependency Graphs}

For each environment $E_i$, we extract entity types and attributes from its initial database and tool descriptions. We then construct a local entity--tool graph $G_i=(V_i,R_i)$, where nodes represent entity types and directed edges represent relations supported by tools. In Figure~\ref{fig:pipeline}, for example, \texttt{reserve\_room} supports a booking relation from \texttt{Person} to \texttt{Hotel}, while \texttt{search\_hotels} supports a search relation from \texttt{City} to \texttt{Hotel}.

\subsection{Cross-Environment Graph Composition}

\noindent\textbf{Graph composition.} Different environments may support different operations involving the same entity types. We merge their local graphs through these shared types, retaining each relation's source environment and supporting tools. In Figure~\ref{fig:pipeline}, shared \texttt{Person} and \texttt{City} nodes connect the flight, hotel, and dining graphs, bringing their operations into a cross-environment entity--tool graph.

\noindent\textbf{Task skeleton sampling.} From the composed graph, we sample connected subgraphs spanning at least two environments. We assess whether the selected operations involve related entities, complement one another, and form a coherent business workflow. For each selected combination, we define a task objective specifying how the operations work together, including which actions use results from other environments and what constraints their combined outcomes must satisfy. This produces a task skeleton $S$ and its required tools. We control task composition through environment count, business-domain scope, and whether user interaction is required.

\subsection{Tool Synthesis and Testing}

Given a task skeleton, we reuse tested tool implementations and have a coding agent implement missing tools using their descriptions, research findings, and initial databases. We test required tools for correct outputs, intended state changes, and error handling, repairing and retesting failures. The tested implementations provide executable operations for the participating environments and support subsequent task instantiation.

\subsection{Task Instantiation and Verification}

\noindent\textbf{Data alignment.} Given a task skeleton and its objective, we select concrete entities and records, supplement missing data, and resolve conflicting information across environments. We plan initial data and reference actions jointly: required records must exist initially or be produced by earlier actions, while records the agent must create are not inserted in advance. This produces the aligned initial state $\mathbf s_{0,\tau}$.

\noindent\textbf{Reference execution.} We execute the planned actions from $\mathbf s_{0,\tau}$, recording outputs and state changes. Actions use results from other environments as required by the objective. Execution failures trigger revisions to the data and action plan. Successful execution provides the reference sequence $A^*$, from which we generate the task instruction $I$ together with the task scenario.

\noindent\textbf{Verification rule generation.} We derive executable rules $C_\tau$ from the task objective and constraints, using reference execution results to specify expected outcomes. The rules check tool outputs, required state changes, and cross-environment consistency. Negative cases, such as omitted operations and incorrect values, test whether the rules reject incomplete or incorrect outcomes.

\noindent\textbf{Task validation.} We reset the environments and replay $A^*$, requiring reproducible outcomes that pass all verification rules. We then run five independent agent attempts from the same initial state, using the task instruction and available interaction interfaces. We retain tasks with successful reference replay and at least two successful agent attempts.

\section{CrossEnv-Curriculum}
\label{sec:crossenv-curriculum}

\noindent\textbf{Curriculum organization.} We organize CrossEnv tasks into five levels according to environment count, business-domain scope, and user interaction. \textbf{L1} contains single-environment tasks that establish basic tool-use skills. \textbf{L2} introduces coordination between two environments within one business domain, and \textbf{L3} expands to three or more environments within the same domain. \textbf{L4} extends coordination across multiple business domains while retaining three or more environments. \textbf{L5} retains L4's environment-count and domain-scope requirements and introduces user interaction. The curriculum progressively expands environment count and business-domain scope, then introduces user interaction under L4's structural requirements.

\noindent\textbf{Training procedure.} We first perform supervised fine-tuning on L1 reference trajectories to teach tool selection, argument specification, and the use of tool outputs. We then conduct reinforcement learning sequentially on L2--L5, initializing each stage from the preceding checkpoint. During RL, the agent acts on the task instruction and interaction history, receiving rewards from executable checks of tool outputs and final environment states.

\section{Experiments}

We present the experimental setup (\S\ref{sec:experimental-setup}) and overall performance (\S\ref{sec:overall}), then analyze curriculum learning (\S\ref{sec:curriculum-analysis}), generalization (\S\ref{sec:generalization}), the value of cross-environment tasks (\S\ref{sec:crossenv-training}), and synthesis quality and construction cost (\S\ref{sec:synthesis-quality-cost}).

\subsection{Experimental Setup}
\label{sec:experimental-setup}

\noindent\textbf{Benchmarks.} Our main evaluation uses four benchmarks covering complementary capabilities: \textbf{interactive tool use}, assessed by $\tau^2$-bench~\citep{barres2025tau2benchevaluatingconversationalagents} through multi-step execution and user coordination; \textbf{long-horizon task completion}, assessed by DeepPlanning-Shopping~\citep{zhang2026deepplanningbenchmarkinglonghorizonagentic} through constrained planning and VitaBench~\citep{he2025vitabenchbenchmarkingllmagents} through coordinated service interactions; and \textbf{general assistant tasks}, assessed by Claw-Eval~\citep{ye2026clawevaltrustworthyevaluationautonomous} through service orchestration and professional dialogue. 
Benchmark subsets and evaluation settings are detailed in Appendix~\ref{app:evaluation-protocol}.

\begin{table}[t]
\centering
\small
\renewcommand{\arraystretch}{1.12}
\setlength{\tabcolsep}{3pt}
\begin{tabular*}{\linewidth}{@{\extracolsep{\fill}}ccrrr@{}}
\toprule
\textbf{Level} & \textbf{Stage} & \textbf{Train} & \textbf{Val.} & \textbf{\shortstack{Avg. Turns}} \\
\midrule
L1 & SFT & 800 & 40 & 4.73 \\
L2 & RL  & 579 & 40 & 4.84 \\
L3 & RL  & 620 & 50 & 8.70 \\
L4 & RL  & 600 & 50 & 12.66 \\
L5 & RL + User & 355 & 40 & 13.76 \\
\bottomrule
\end{tabular*}
\caption{CrossEnv training and validation data by curriculum level. Turns count tool calls and, for L5, one required clarification question.}
\label{tab:training-data}
\end{table}

\begin{table*}[!t]
\centering
\fontsize{7.5}{8.5}\selectfont
\setlength{\tabcolsep}{0.2pt}
\renewcommand{\arraystretch}{1.25}
\newcommand{\traindata}[1]{\makebox[\linewidth][r]{#1}}
\newcommand{\institutionmark}[2][2.3mm]{%
  \raisebox{\dimexpr 2.3pt-\height/2\relax}{\includegraphics[width=#1,height=2.3mm,keepaspectratio]{figures/table3_institution_logos_v1/#2}}%
}
\newcommand{\methodmark}[3][2.3mm]{%
  \makebox[4mm][c]{\institutionmark[#1]{#2}}\hspace{0.6mm}#3%
}
\newcommand{\crossenvmethod}[1]{%
  \makebox[4mm][c]{\raisebox{\dimexpr 2.3pt-\height/2\relax}{\includegraphics[width=2.6mm]{figures/crossenv_logo_v3/crossenv_icon.pdf}}}\hspace{0.6mm}#1%
}
\begin{tabularx}{\textwidth}{@{}p{25.3mm}@{\hspace{0.6mm}}>{\raggedleft\arraybackslash}p{9mm}@{\hspace{0.8mm}}>{\centering\arraybackslash}p{6.5mm}@{\hspace{2.0mm}}>{\raggedleft\arraybackslash}X>{\raggedleft\arraybackslash}X>{\raggedleft\arraybackslash}X>{\raggedleft\arraybackslash}X>{\columncolor{black!4}\raggedleft\arraybackslash}X!{\hspace{1.2mm}}>{\raggedleft\arraybackslash}X>{\raggedleft\arraybackslash}X>{\raggedleft\arraybackslash}X>{\columncolor{black!4}\raggedleft\arraybackslash}X!{\hspace{1.2mm}}>{\raggedleft\arraybackslash}X>{\raggedleft\arraybackslash}X>{\raggedleft\arraybackslash}X>{\raggedleft\arraybackslash}X>{\columncolor{black!4}\raggedleft\arraybackslash}X!{\hspace{1.2mm}}>{\raggedleft\arraybackslash}X>{\raggedleft\arraybackslash}X>{\columncolor{black!4}\raggedleft\arraybackslash}X@{}}
\toprule
\multicolumn{1}{@{}l}{\multirow{2}{*}[-2pt]{\textbf{Method}}} & \multicolumn{1}{c}{\multirow{2}{*}[-2pt]{\textbf{\shortstack[c]{Train\\data}}}} & \multicolumn{1}{c}{\multirow{2}{*}[-2pt]{\textbf{Source}}} & \multicolumn{5}{c}{\textbf{$\tau^2$-bench}} & \multicolumn{4}{c}{\textbf{DeepPlanning}} & \multicolumn{5}{c}{\textbf{VitaBench}} & \multicolumn{3}{c}{\textbf{Claw-Eval}} \\
\cmidrule(lr){4-8}\cmidrule(lr){9-12}\cmidrule(lr){13-17}\cmidrule(lr){18-20}
 & & & Ret. & Air. & Tel. & Bnk. & \textbf{Avg.} & L1 & L2 & L3 & \textbf{Avg.} & Del. & Ins. & OTA & Cr. & \textbf{Avg.} & Gen. & MT & \textbf{Avg.} \\
\midrule
\rowcolor{black!4}
\multicolumn{20}{c}{\textit{Frontier Closed-Source Models}} \\
\methodmark{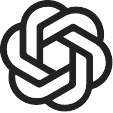}{GPT-5.6-Luna} & \multicolumn{1}{c}{---} & \sourceAPI & 21.9 & 44.0 & 9.6 & 6.2 & 17.1 & 40.0 & 40.0 & 20.0 & 36.7 & 32.0 & 40.0 & 8.0 & 12.0 & 23.0 & 78.3 & 57.9 & 74.4 \\
\methodmark{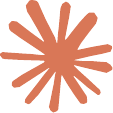}{Claude-Haiku-4.5} & \multicolumn{1}{c}{---} & \sourceAPI & 36.0 & 58.0 & 1.8 & 3.1 & 20.0 & 36.0 & 24.0 & 15.0 & 27.5 & 40.0 & 32.0 & 24.0 & 8.0 & 26.0 & 49.0 & 36.8 & 46.7 \\
\methodmark{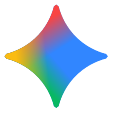}{Gemini-3.8-Flash} & \multicolumn{1}{c}{---} & \sourceAPI & 9.6 & 24.0 & 9.6 & 2.1 & 9.6 & 34.0 & 34.0 & 25.0 & 32.5 & 0.0 & 0.0 & 48.0 & 0.0 & 12.0 & 68.8 & 84.2 & 71.8 \\
\methodmark{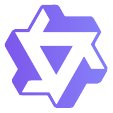}{Qwen3.8-Max} & \multicolumn{1}{c}{---} & \sourceAPI & 30.7 & 50.0 & 7.9 & 4.1 & 19.5 & 72.0 & 66.0 & 75.0 & 70.0 & 64.0 & 52.0 & 36.0 & 36.0 & 47.0 & 78.3 & 76.3 & 77.9 \\
\midrule
\rowcolor{black!4}
\multicolumn{20}{c}{\textit{Open-Weight General-Purpose Models}} \\
\methodmark{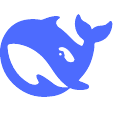}{DeepSeek-V4-Pro} & \multicolumn{1}{c}{---} & \sourceAPI & 43.9 & 62.0 & 7.0 & 4.1 & 24.8 & 62.0 & 38.0 & 50.0 & 50.0 & 44.0 & 48.0 & 36.0 & 28.0 & 39.0 & 75.2 & 81.6 & 76.4 \\
\methodmark{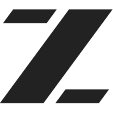}{GLM-5.3} & \multicolumn{1}{c}{---} & \sourceAPI & 16.7 & 36.0 & 4.4 & 3.1 & 12.0 & 62.0 & 42.0 & 30.0 & 48.3 & 64.0 & 64.0 & 60.0 & 40.0 & 57.0 & 77.1 & 71.1 & 75.9 \\
\methodmark{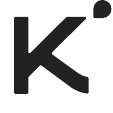}{Kimi-K3} & \multicolumn{1}{c}{---} & \sourceAPI & 20.2 & 42.0 & 3.5 & 4.1 & 13.9 & 42.0 & 24.0 & 25.0 & 31.7 & 60.0 & 48.0 & 44.0 & 32.0 & 46.0 & 80.9 & 78.9 & 80.5 \\
\methodmark{qwen.pdf}{Qwen3-1.7B} & \multicolumn{1}{c}{---} & \sourcePublic & 5.3 & 14.0 & 4.4 & 3.1 & 5.6 & 4.0 & 0.0 & 0.0 & 1.7 & 0.0 & 0.0 & 0.0 & 0.0 & 0.0 & 8.9 & 2.6 & 7.7 \\
\methodmark{qwen.pdf}{Qwen3-4B} & \multicolumn{1}{c}{---} & \sourcePublic & 9.6 & \underline{28.0} & 5.3 & 2.1 & 8.8 & 2.0 & 0.0 & 0.0 & 0.8 & 0.0 & 4.0 & 0.0 & 0.0 & 1.0 & 15.3 & 0.0 & 12.3 \\
\methodmark{qwen.pdf}{Qwen3-8B} & \multicolumn{1}{c}{---} & \sourcePublic & \underline{14.9} & 26.0 & 7.0 & \textbf{5.2} & 11.5 & 4.0 & 0.0 & 0.0 & 1.7 & 16.0 & \underline{12.0} & 0.0 & 0.0 & 7.0 & 17.2 & 2.6 & 14.4 \\
\midrule
\rowcolor{black!4}
\multicolumn{20}{c}{\textit{Environment-Trained Baselines}} \\
\methodmark{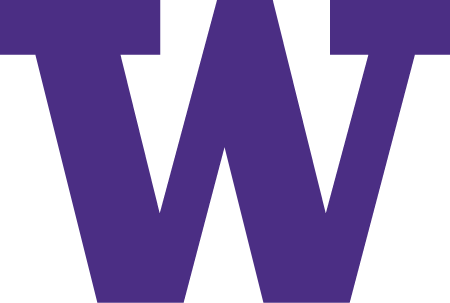}{Toucan-7B} & \traindata{119,300} & \sourcePublic & 9.6 & 8.0 & 0.0 & 3.1 & 4.8 & 6.0 & 2.0 & 0.0 & 3.3 & 0.0 & 0.0 & 0.0 & 0.0 & 0.0 & \underline{26.1} & 2.6 & \underline{21.5} \\
\methodmark{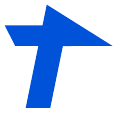}{EnvACE-1.7B} & \traindata{8,129} & \sourcePublic & 4.4 & 18.0 & 0.9 & 2.1 & 4.5 & 2.0 & 0.0 & 0.0 & 0.8 & 0.0 & 0.0 & 0.0 & 0.0 & 0.0 & 6.4 & 0.0 & 5.1 \\
\methodmark{tencent-color.pdf}{EnvACE-8B} & \traindata{8,129} & \sourcePublic & 12.3 & \textbf{30.0} & 3.5 & 3.1 & 9.6 & 4.0 & 2.0 & 0.0 & 2.5 & 8.0 & 0.0 & 0.0 & 0.0 & 2.0 & 16.6 & \textbf{7.9} & 14.9 \\
\methodmark{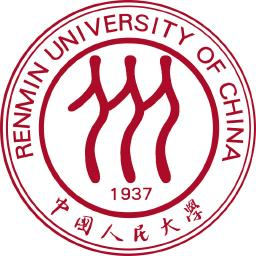}{EnvScaler-1.7B} & \traindata{11,572} & \sourcePublic & 7.9 & 10.0 & 0.9 & 1.0 & 4.3 & 6.0 & 4.0 & 0.0 & 4.2 & 8.0 & 0.0 & 0.0 & 0.0 & 2.0 & 11.5 & 2.6 & 9.7 \\
\methodmark{ruc_mark.png}{EnvScaler-4B} & \traindata{11,572} & \sourceInhouse & 11.4 & 26.0 & \textbf{10.5} & \textbf{5.2} & 11.5 & 8.0 & 4.0 & \textbf{10.0} & 6.7 & \underline{24.0} & 0.0 & 0.0 & 0.0 & 6.0 & 13.4 & 0.0 & 10.8 \\
\methodmark{ruc_mark.png}{EnvScaler-8B} & \traindata{11,572} & \sourcePublic & \textbf{21.9} & 22.0 & 7.0 & \underline{4.1} & \textbf{12.8} & \underline{12.0} & \underline{6.0} & \underline{5.0} & 8.3 & \underline{24.0} & 4.0 & 4.0 & \underline{4.0} & \underline{9.0} & 24.8 & \textbf{7.9} & \underline{21.5} \\
\methodmark{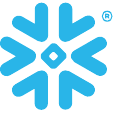}{Arctic-AWM-4B} & \traindata{3,315} & \sourcePublic & 3.5 & 16.0 & 6.1 & 3.1 & 5.9 & 0.0 & 0.0 & 0.0 & 0.0 & 8.0 & 0.0 & 0.0 & 0.0 & 2.0 & 8.3 & 0.0 & 6.7 \\
\methodmark{snowflake.pdf}{Arctic-AWM-8B} & \traindata{3,315} & \sourcePublic & 12.3 & 26.0 & \textbf{10.5} & \underline{4.1} & 11.5 & 2.0 & 2.0 & 0.0 & 1.7 & 8.0 & \underline{12.0} & 0.0 & 0.0 & 5.0 & 15.9 & \textbf{7.9} & 14.4 \\
\methodmark{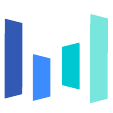}{Agent-World-4B} & \traindata{3,000} & \sourceInhouse & 12.3 & 26.0 & 7.0 & 1.0 & 9.6 & \underline{12.0} & 4.0 & 0.0 & 6.7 & \underline{24.0} & 4.0 & 0.0 & 0.0 & 7.0 & 21.7 & 0.0 & 17.4 \\
\methodmark{bytedance.pdf}{Agent-World-8B} & \traindata{3,000} & \sourceInhouse & 14.0 & 18.0 & 7.0 & 2.1 & 9.3 & \textbf{14.0} & \underline{6.0} & \underline{5.0} & \underline{9.2} & \textbf{36.0} & 8.0 & \textbf{12.0} & 0.0 & \textbf{14.0} & 24.2 & \underline{5.3} & 20.5 \\
\midrule
\rowcolor{black!4}
\multicolumn{20}{c}{\textit{Our Models}} \\
\crossenvmethod{\textbf{CrossEnv-1.7B}} & \traindata{2,954} & \sourceInhouse & 5.3 & 12.0 & 7.9 & 3.1 & 6.4 & \underline{12.0} & 4.0 & 0.0 & 6.7 & 4.0 & 4.0 & 0.0 & 0.0 & 2.0 & 8.9 & 2.6 & 7.7 \\
\crossenvmethod{\textbf{CrossEnv-4B}} & \traindata{2,954} & \sourceInhouse & 11.4 & 22.0 & \textbf{10.5} & 2.1 & 10.1 & \underline{12.0} & \textbf{10.0} & 0.0 & \underline{9.2} & 8.0 & 8.0 & \underline{8.0} & \underline{4.0} & 7.0 & 22.3 & \underline{5.3} & 19.0 \\
\crossenvmethod{\textbf{CrossEnv-8B}} & \traindata{2,954} & \sourceInhouse & \underline{14.9} & \underline{28.0} & \underline{9.7} & \underline{4.1} & \underline{12.3} & \textbf{14.0} & \textbf{10.0} & \underline{5.0} & \textbf{10.8} & \underline{24.0} & \textbf{16.0} & \underline{8.0} & \textbf{8.0} & \textbf{14.0} & \textbf{27.4} & \textbf{7.9} & \textbf{23.6} \\
\bottomrule
\end{tabularx}

\normalsize
\caption[Results on four external benchmarks]{Performance on four external benchmarks. Source: \sourceAPI\ hosted API; \sourcePublic\ public checkpoint deployed locally; \sourceInhouse\ trained in this work. Avg. is weighted by subset size. Bold and underlined scores indicate the best and second-best among the displayed 1.7B--8B models.}
\label{tab:overall-no-bfcl-mcp}
\end{table*}

\noindent\textbf{Models.} We use Qwen3 backbones~\citep{yang2025qwen3technicalreport} with 1.7B, 4B, and 8B parameters and compare against the corresponding base models to assess training gains. Environment-trained baselines include Toucan~\citep{xu2025toucansynthesizing15mtoolagentic}, EnvACE~\citep{xu2026envaceinternalizingenvironmentdynamics}, EnvScaler~\citep{song2026envscalerscalingtoolinteractiveenvironments}, Arctic-AWM~\citep{wang2026agentworldmodelinfinity}, and Agent-World~\citep{dong2026agentworldscalingrealworldenvironment}. We use in-house reproductions of EnvScaler-4B and Agent-World-4B/8B; the Agent-World implementation was not publicly available at the time of our experiments. We also include frontier closed-source models~\citep{openai_gpt56luna,anthropic2025haiku45,google2026gemini38flash,qwen38modelcard} and open-weight general-purpose models~\citep{deepseek2026v4pro0813,zai_glm53,kimiteam2026kimik3openfrontier} as reference points. Model versions and sources are documented in Appendix~\ref{app:model-versions}.

\noindent\textbf{Training.} We train on CrossEnv-generated data following the curriculum in Section~\ref{sec:crossenv-curriculum}. We first perform full-parameter SFT on 800 L1 reference trajectories, followed by stagewise trajectory-level outcome GRPO on 2,154 tasks across L2--L5, for a total of 2,954 training records. Table~\ref{tab:training-data} reports training and validation counts and average reference lengths by level. All CrossEnv models in the main results table complete L1--L5 training. Capability tracking extends through L5; the Mixed comparison and longer-workflow tests use checkpoints through L4.

\subsection{Overall Performance}
\label{sec:overall}

Table~\ref{tab:overall-no-bfcl-mcp} reports results on the four external benchmarks. We highlight three findings.

\noindent\textbf{(1) Consistent gains over base models.} Across all three model sizes, CrossEnv improves or matches the corresponding Qwen3 backbone on each benchmark. At 8B, scores increase from 1.7 to 10.8 on DeepPlanning, 7.0 to 14.0 on VitaBench, and 14.4 to 23.6 on Claw-Eval, showing gains across constrained planning, coordinated service interactions, and general assistant tasks.

\noindent\textbf{(2) Competitive performance with fewer training records.} Among the evaluated 1.7B--8B models, CrossEnv-8B achieves the highest aggregate scores on DeepPlanning and Claw-Eval and ties with Agent-World-8B on VitaBench. With 2,954 training records, approximately one quarter of EnvScaler's reported dataset, it outperforms EnvScaler-8B on three of the four benchmarks.

\noindent\textbf{(3) Gains vary across capabilities and model scales.} CrossEnv improves planning and service coordination across all three model sizes, with larger models also showing stronger gains on general assistant tasks. Improvements on interactive tool use are more modest. These results indicate benefits across multiple capabilities, with the extent of improvement varying by task type and model scale.

\begin{figure*}[!t]
\centering
\includegraphics[width=\linewidth]{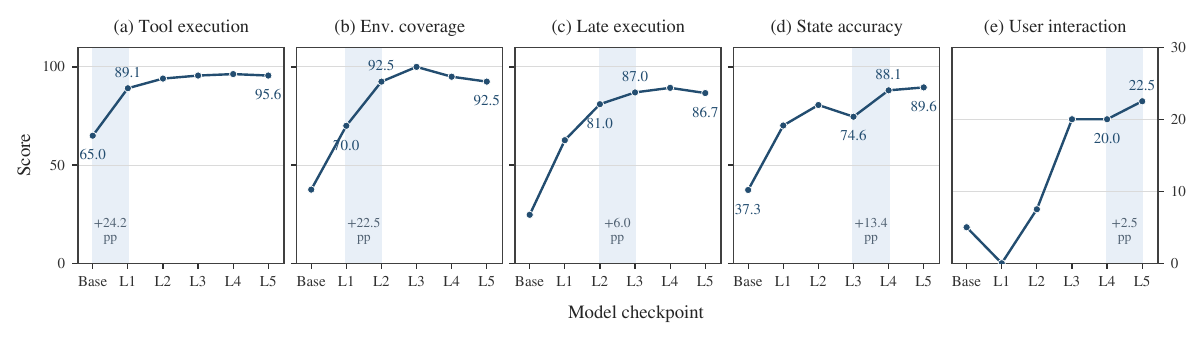}
\caption{Qwen3-8B capability scores from Base to L5. Shading highlights the corresponding training stages. Panel (e) reports a composite interaction score, not a success rate (Appendix~\ref{app:user-interaction-metric}).}
\label{fig:curriculum-analysis}
\end{figure*}

\begin{figure}[!t]
\centering
\includegraphics[width=\linewidth]{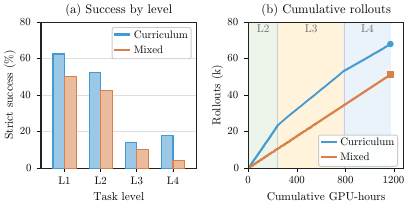}
\caption{Curriculum vs. Mixed (8B): (a) strict success at similar GPU-hours; (b) cumulative rollouts.}
\label{fig:curriculum-mixed}
\end{figure}

\subsection{Curriculum Learning Analysis}
\label{sec:curriculum-analysis}

\noindent\textbf{Capability development and retention.}
We track Qwen3-8B from Base through L5 on fixed evaluation sets (Figure~\ref{fig:curriculum-analysis}). L1 SFT improves basic tool use, while subsequent stages improve the model's ability to act across required environments and execute later workflow steps. These gains do not always translate immediately into correct outcomes: the model executes more required tools at L3 but passes fewer final-state checks. More final-state checks pass at L4 and L5, with user interaction also improving at L5. Although some execution scores decrease slightly at later stages, most earlier gains are retained. Overall, the model learns to coordinate more complex workflows while largely preserving previously acquired tool-use skills.

\begin{table}[!t]
\centering
\fontsize{9}{11}\selectfont
\setlength{\tabcolsep}{2.5pt}
\renewcommand{\arraystretch}{1.12}
\begin{tabularx}{\linewidth}{@{}ll*{4}{>{\raggedleft\arraybackslash}X}@{}}
\toprule
\textbf{Model} & \textbf{Pair} & \textbf{Success} & \textbf{Obs.} & \textbf{State} & \textbf{Answer} \\
\midrule
\multirow{2}{*}{\textbf{1.7B}} & Seen & +6.7 & +40.7 & +37.5 & +22.2 \\
 & \cellcolor{black!4}\textbf{Unseen} & \cellcolor{black!4}+5.3 & \cellcolor{black!4}+38.9 & \cellcolor{black!4}\phantom{+}0.0 & \cellcolor{black!4}+11.1 \\
\midrule
\multirow{2}{*}{\textbf{4B}} & Seen & +30.0 & +43.7 & +48.5 & +32.2 \\
 & \cellcolor{black!4}\textbf{Unseen} & \cellcolor{black!4}+24.0 & \cellcolor{black!4}+33.7 & \cellcolor{black!4}+26.9 & \cellcolor{black!4}+22.5 \\
\midrule
\multirow{2}{*}{\textbf{8B}} & Seen & +28.0 & +44.3 & +30.4 & +26.4 \\
 & \cellcolor{black!4}\textbf{Unseen} & \cellcolor{black!4}+16.0 & \cellcolor{black!4}+34.3 & \cellcolor{black!4}+14.6 & \cellcolor{black!4}+13.0 \\
\bottomrule
\end{tabularx}
\caption{CrossEnv gains over Base on seen and unseen environment combinations (percentage points). Obs.: observation.}
\label{tab:generalization-combinations}
\end{table}

\noindent\textbf{Comparison with mixed training.} We compare curriculum and mixed training from the same L1 SFT checkpoint using the same L2--L4 task pools. At checkpoints with similar cumulative RL GPU-hours, curriculum training achieves higher success at every task level (Figure~\ref{fig:curriculum-mixed}(a)). Starting with shorter two-environment tasks before introducing longer, more complex workflows may reduce the interaction cost per sample early in training. Curriculum training also accumulates more rollouts within a comparable compute budget (Figure~\ref{fig:curriculum-mixed}(b)). These results support the value of the progressive curriculum for both training effectiveness and efficient use of compute under the current configurations.

\subsection{Generalization Analysis}
\label{sec:generalization}

We examine whether learned skills transfer to new environment combinations, longer workflows, and external tasks less similar to training data.

\noindent\textbf{Generalization to unseen environment combinations.} We evaluate combinations of familiar environments that were never used together in a training task. Table~\ref{tab:generalization-combinations} compares gains over Base on seen and unseen combinations, with task success requiring all task requirements to be met. CrossEnv improves task success on both groups at all three model sizes, showing transfer to new combinations of familiar environments.

\begin{figure}[!t]
\centering
\includegraphics[width=\linewidth]{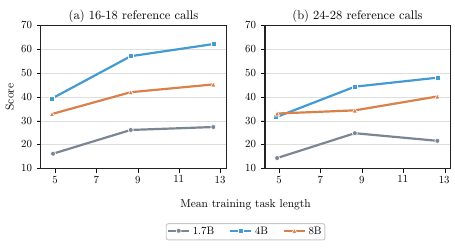}
\caption{Generalization to longer workflows. Scores show the average proportion of task checks passed (\%).}
\label{fig:long-horizon-generalization}
\end{figure}

\noindent\textbf{Generalization to longer workflows.} We evaluate curriculum checkpoints on two sets of tasks with reference workflows of 16--18 and 24--28 tool calls, both exceeding the training maximum of 14 calls. Figure~\ref{fig:long-horizon-generalization} shows the average proportion of task checks passed. From L1 to L4, both 4B and 8B improve on both sets, while 1.7B shows a late-stage decline on longer tasks. These results suggest that training on shorter workflows can improve performance on longer tasks, although gains vary by model size.

\begin{figure}[!t]
\centering
\includegraphics[width=\linewidth]{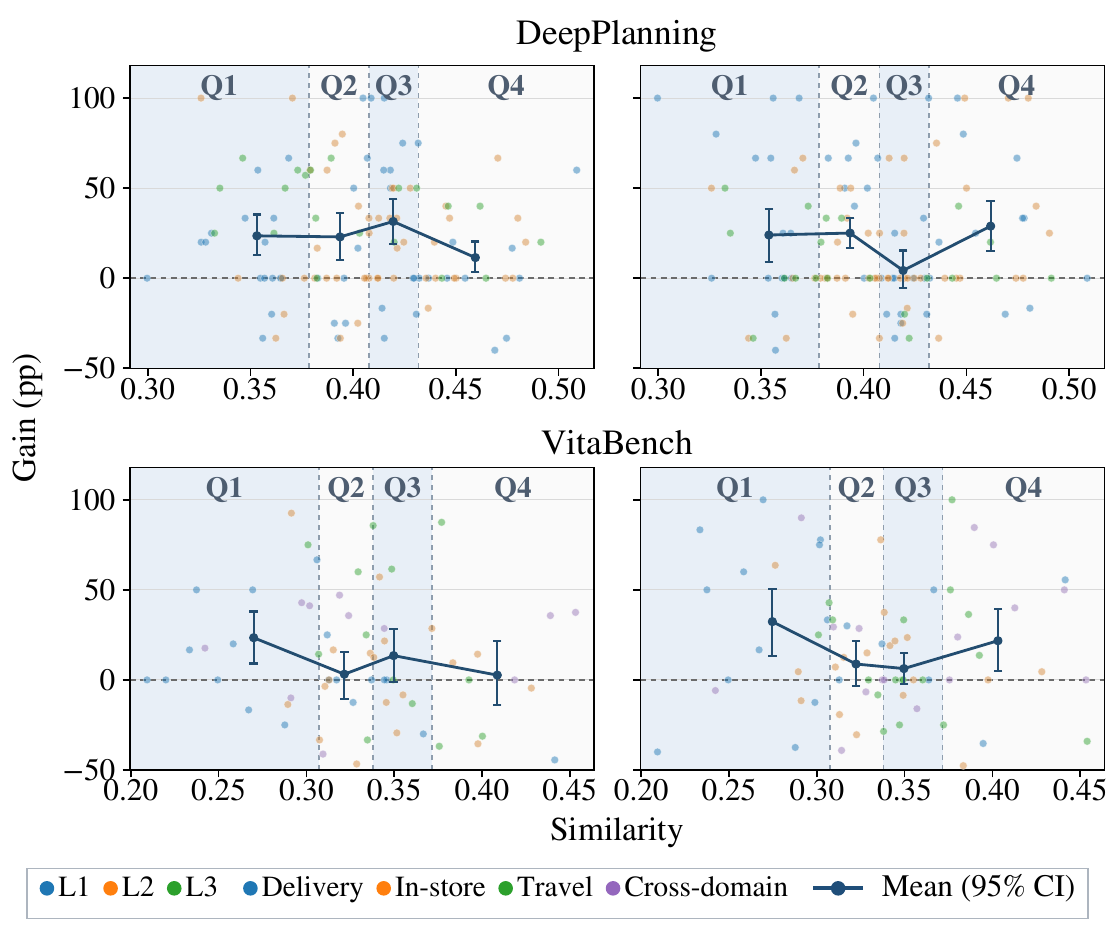}
\caption{Task similarity and CrossEnv gains over Base at 4B (left) and 8B (right). Lines show mean gains in four similarity groups with 95\% confidence intervals.}
\label{fig:vita-similarity}
\end{figure}

\noindent\textbf{Transfer across task similarity.} We examine whether CrossEnv learns transferable skills beyond its training tasks. Inspired by EnvScaler~\citep{song2026envscalerscalingtoolinteractiveenvironments}, we use instruction-level semantic similarity as a proxy for task similarity. In Figure~\ref{fig:vita-similarity}, both model sizes improve over Base on average in the lowest-similarity group of each benchmark, while higher similarity does not consistently yield larger gains. These results support generalization to tasks less similar to the training data.

\subsection{The Value of Cross-Environment Tasks}
\label{sec:crossenv-training}

\begin{table}[!t]
\centering
\fontsize{8.5}{10}\selectfont
\setlength{\tabcolsep}{2pt}
\renewcommand{\arraystretch}{1.16}
\begin{tabular*}{\linewidth}{@{\extracolsep{\fill}}lrrr>{\columncolor{black!4}[\tabcolsep][0pt]}r@{}}
\toprule
\textbf{Metric} & \textbf{Base} & \textbf{SFT} & \textbf{SingleEnv} & \textbf{CrossEnv} \\
\midrule
\multicolumn{5}{@{}c@{}}{\textit{Benchmark performance}} \\
\midrule
$\tau^2$-bench & \textbf{11.5} & 8.5 & 8.8 & \textbf{11.5} \\
DeepPlanning & 1.7 & 3.3 & 6.7 & \textbf{7.5} \\
Claw-Eval & 14.4 & 16.9 & 15.4 & \textbf{20.0} \\
\midrule
\multicolumn{5}{@{}c@{}}{\textit{Execution capabilities}} \\
\midrule
Output reuse & 6.7 & 8.3 & 1.7 & \textbf{11.7} \\
Later-step success & 23.3 & 18.3 & 28.3 & \textbf{33.3} \\
Joint success & 5.0 & 1.7 & 0.0 & \textbf{8.3} \\
\bottomrule
\end{tabular*}
\caption{Qwen3-8B comparison (\%). Bold: row best.}
\label{tab:crossenv-training-performance}
\end{table}

\noindent\textbf{Performance comparison.} We compare single- and cross-environment RL using the same SFT cold-start initialization (Table~\ref{tab:crossenv-training-performance}). CrossEnv outperforms Base, SFT, and SingleEnv on DeepPlanning and Claw-Eval; on $\tau^2$-bench, it exceeds SFT and SingleEnv and matches Base.

\noindent\textbf{Execution analysis.} To understand why cross-environment training yields larger gains than single-environment training, we systematically analyze evaluation trajectories across the four evaluation benchmarks. The differences center on three aspects: (1) \textbf{output reuse}, correctly using prior tool outputs in subsequent operations; (2) \textbf{later-step success}, continuing execution to complete required later operations; and (3) \textbf{joint success}, satisfying all service goals and task constraints together (Appendix~\ref{app:crossenv-value}). CrossEnv outperforms SingleEnv on all three metrics (Table~\ref{tab:crossenv-training-performance}), suggesting that cross-environment training helps models coordinate multiple operations toward a shared goal.

\subsection{Synthesis Quality and Construction Cost}
\label{sec:synthesis-quality-cost}

\begin{figure}[!t]
\centering
\includegraphics[width=\columnwidth]{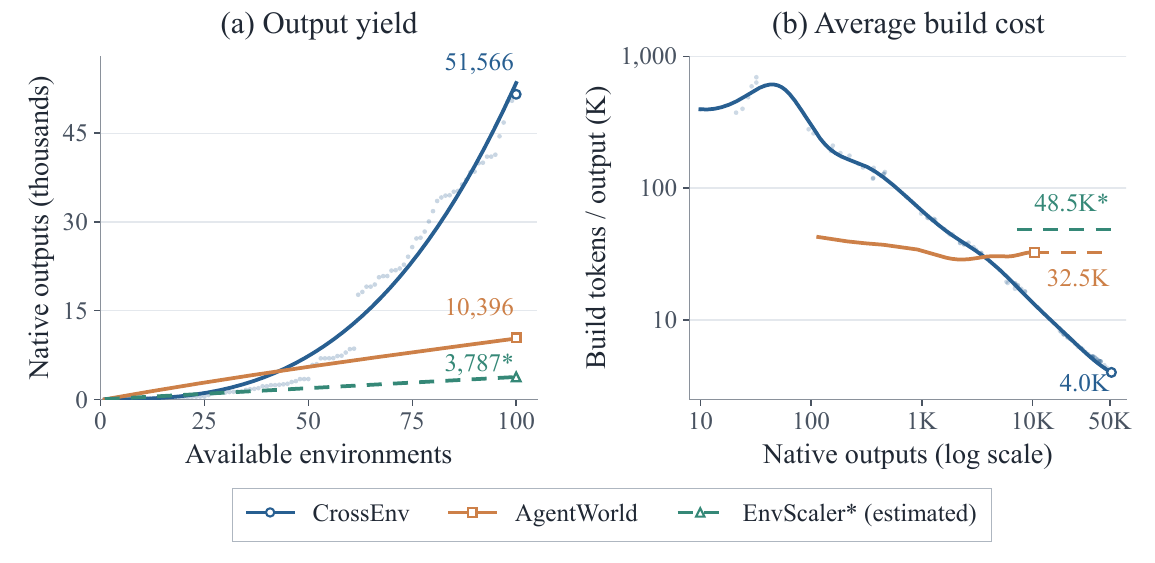}
\caption{Output yield and average environment-build cost. *: estimates.}
\label{fig:task-yield-build-cost}
\end{figure}

\noindent\textbf{Execution and verification quality.} We randomly sample 100 tasks, stratified by domain, with 50 each from MCP and Skill environments. All reference workflows run successfully and pass all checks. Further tests on subsets confirm that: (1) resetting the environments and rerunning the workflows yields the same results; (2) skipping a required operation prevents full credit; and (3) reordering queries without changing their results still earns full credit (Appendix~\ref{app:synthesis-quality}).

\noindent\textbf{Scalability and construction cost.} As the environment pool grows, CrossEnv combines new environments with existing ones, enabling more task combinations and accelerating growth in candidate task yield (Figure~\ref{fig:task-yield-build-cost}(a)). These combinations reuse environment data, tools, and entity relations without repeating environment construction, so the average construction cost per candidate generally decreases as output grows (Figure~\ref{fig:task-yield-build-cost}(b)). Although supporting long cross-environment workflows requires a higher initial investment, this reuse eventually reduces CrossEnv's average construction cost to approximately 4.0k tokens per candidate, below our reproduced AgentWorld result and the EnvScaler estimate based on reported figures.

\section{Conclusion}
\label{sec:conclusion}

We presented CrossEnv, a framework for synthesizing executable and verifiable cross-environment tasks with consistent data, and CrossEnv-Curriculum, a progressive training scheme for learning from these tasks. Across four external benchmarks, combining cross-environment data with curriculum training achieves competitive performance with fewer training examples. Our analyses suggest that cross-environment tasks improve workflow coordination, while progressive training achieves higher success and collects more rollouts at comparable compute. Further evaluations provide evidence of transfer to new environment combinations, external environments, and workflows longer than those seen during training.

\section*{Limitations}

Programmatically synthesized environments may simplify real-world service behavior, including changing interfaces, noisy responses, and complex business constraints. Although CrossEnv aligns task-specific data to support coherent workflows, transfer to real services with dynamically changing data and operational conditions requires further evaluation.

Our generalization evidence is bounded by the evaluated environments and workflow lengths. The combination experiments test new combinations of familiar environments, while the longer-workflow experiments extend to 24--28 reference tool calls and measure the proportion of task checks passed. These results do not establish reliable end-to-end completion in entirely new services or substantially longer workflows.

Some training comparisons do not fully isolate the factors contributing to performance gains. In particular, curriculum and mixed training use similar cumulative GPU-hours but differ in batch configuration, parallelism, and reference-model settings, limiting attribution of the gains solely to curriculum order. Likewise, comparisons of training-record counts across methods do not establish equivalent advantages in training compute.

\section*{Ethical Considerations}

CrossEnv aims to reduce the effort required to construct cross-environment tasks for agent training. By synthesizing executable and verifiable workflows, it supports research on tool use and service coordination. This capability has potential dual-use implications: tasks designed for beneficial automation could also be adapted to train agents for harmful purposes. We encourage responsible use of the framework and synthesized tasks, and careful evaluation of trained agents before deployment in sensitive applications.

\subsection*{AI Use Statement}
We used generative AI tools to assist with manuscript drafting and language editing. The authors reviewed the resulting content and take full responsibility for the final manuscript.

\FloatBarrier
\bibliography{custom}

\begin{thebibliography}{53}
\providecommand{\natexlab}[1]{#1}

\bibitem[{Acikgoz et~al.(2026)Acikgoz, Qian, H{\"u}botter, Ji, Hakkani-T{\"u}r, and Tur}]{acikgoz2026toolr0selfevolvingllmagents}
Emre~Can Acikgoz, Cheng Qian, Jonas H{\"u}botter, Heng Ji, Dilek Hakkani-T{\"u}r, and Gokhan Tur. 2026.
\newblock \href {https://arxiv.org/abs/2602.21320} {{Tool-R0: Self-Evolving LLM Agents for Tool-Learning from Zero Data}}.
\newblock \emph{Preprint}, arXiv:2602.21320.

\bibitem[{{Anthropic}(2025)}]{anthropic2025haiku45}
{Anthropic}. 2025.
\newblock \href {https://www.anthropic.com/news/claude-haiku-4-5} {{Introducing Claude Haiku 4.5}}.
\newblock Accessed September 21, 2026.

\bibitem[{Barres et~al.(2025)Barres, Dong, Ray, Si, and Narasimhan}]{barres2025tau2benchevaluatingconversationalagents}
Victor Barres, Honghua Dong, Soham Ray, Xujie Si, and Karthik Narasimhan. 2025.
\newblock \href {https://arxiv.org/abs/2506.07982} {{$\tau^2$-Bench: Evaluating Conversational Agents in a Dual-Control Environment}}.
\newblock \emph{Preprint}, arXiv:2506.07982.

\bibitem[{Bengio et~al.(2009)Bengio, Louradour, Collobert, and Weston}]{Bengio_2009}
Yoshua Bengio, J{\'e}r{\^o}me Louradour, Ronan Collobert, and Jason Weston. 2009.
\newblock \href {https://doi.org/10.1145/1553374.1553380} {{Curriculum learning}}.
\newblock In \emph{{Proceedings of the 26th Annual International Conference on Machine Learning}}, ICML '09, pages 41--48. ACM.

\bibitem[{Bu et~al.(2025)Bu, Wu, Yu, Gao, Miao, Zhang, Pan, Li, Li, Ji, Li, Tang, and Zhuang}]{pmlr-v267-bu25b}
Wendong Bu, Yang Wu, Qifan Yu, Minghe Gao, Bingchen Miao, Zhenkui Zhang, Kaihang Pan, Yunfei Li, Mengze Li, Wei Ji, Juncheng Li, Siliang Tang, and Yueting Zhuang. 2025.
\newblock \href {https://proceedings.mlr.press/v267/bu25b.html} {{What Limits Virtual Agent Application? {O}mni{B}ench: A Scalable Multi-Dimensional Benchmark for Essential Virtual Agent Capabilities}}.
\newblock In \emph{Proceedings of the 42nd International Conference on Machine Learning}, volume 267 of \emph{Proceedings of Machine Learning Research}, pages 5725--5748. PMLR.

\bibitem[{Chen et~al.(2025)Chen, Zhao, Zhang, Liu, Qi, Wu, Kalluri, Cao, Xiong, Tong, Yao, Li, Zhu, Li, Song, Li, Weston, and Huynh}]{chen2025scalingagentlearningexperience}
Zhaorun Chen, Zhuokai Zhao, Kai Zhang, Bo~Liu, Qi~Qi, Yifan Wu, Tarun Kalluri, Sara Cao, Yuanhao Xiong, Haibo Tong, Huaxiu Yao, Hengduo Li, Jiacheng Zhu, Xian Li, Dawn Song, Bo~Li, Jason Weston, and Dat Huynh. 2025.
\newblock \href {https://arxiv.org/abs/2511.03773} {{Scaling Agent Learning via Experience Synthesis}}.
\newblock \emph{Preprint}, arXiv:2511.03773.

\bibitem[{{DeepSeek}(2026)}]{deepseek2026v4pro0813}
{DeepSeek}. 2026.
\newblock \href {https://api-docs.deepseek.com/updates/#date-2026-08-13} {{DeepSeek-V4-Pro Update}}.
\newblock Accessed September 21, 2026.

\bibitem[{Dong et~al.(2026)Dong, Lu, Huang, Zhong, Liu, Huang, Li, Zhao, Song, Li, Jin, Zhu, Wang, Lei, Luo, Chen, Chen, Feng, Wen, and Dou}]{dong2026agentworldscalingrealworldenvironment}
Guanting Dong, Junting Lu, Junjie Huang, Wanjun Zhong, Longxiang Liu, Shijue Huang, Zhenyu Li, Yang Zhao, Xiaoshuai Song, Xiaoxi Li, Jiajie Jin, Yutao Zhu, Hanbin Wang, Fangyu Lei, Qinyu Luo, Mingyang Chen, Zehui Chen, Jiazhan Feng, Ji-Rong Wen, and Zhicheng Dou. 2026.
\newblock \href {https://arxiv.org/abs/2604.18292} {{Agent-World: Scaling Real-World Environment Synthesis for Evolving General Agent Intelligence}}.
\newblock \emph{Preprint}, arXiv:2604.18292.

\bibitem[{Fan et~al.(2026)Fan, Yu, Cai, Guan, Yang, Hu, Zhou, Wu, Han, Zhang, and Wang}]{fan2026scalableterminaltasksynthesis}
Zhiyuan Fan, Tinghao Yu, Yuanjun Cai, Jiangtao Guan, Yun Yang, Dingxin Hu, Jiang Zhou, Xing Wu, Zhuo Han, Feng Zhang, and Lilin Wang. 2026.
\newblock \href {https://arxiv.org/abs/2604.25727} {{Toward Scalable Terminal Task Synthesis via Skill Graphs}}.
\newblock \emph{Preprint}, arXiv:2604.25727.

\bibitem[{Florensa et~al.(2018)Florensa, Held, Geng, and Abbeel}]{pmlr-v80-florensa18a}
Carlos Florensa, David Held, Xinyang Geng, and Pieter Abbeel. 2018.
\newblock \href {https://proceedings.mlr.press/v80/florensa18a.html} {{Automatic Goal Generation for Reinforcement Learning Agents}}.
\newblock In \emph{{Proceedings of the 35th International Conference on Machine Learning}}, volume~80 of \emph{Proceedings of Machine Learning Research}, pages 1515--1528. PMLR.

\bibitem[{Florensa et~al.(2017)Florensa, Held, Wulfmeier, Zhang, and Abbeel}]{pmlr-v78-florensa17a}
Carlos Florensa, David Held, Markus Wulfmeier, Michael Zhang, and Pieter Abbeel. 2017.
\newblock \href {https://proceedings.mlr.press/v78/florensa17a.html} {{Reverse Curriculum Generation for Reinforcement Learning}}.
\newblock In \emph{{Proceedings of the 1st Annual Conference on Robot Learning}}, volume~78 of \emph{Proceedings of Machine Learning Research}, pages 482--495. PMLR.

\bibitem[{Froger et~al.(2025)Froger, Andrews, Bettini, Budhiraja, Cabral, Do, Garreau, Gaya, Lauren{\c c}on, Lecanu, Malkan, Mekala, M{\'e}nard, Bertran, Piterbarg, Plekhanov, Rita, Rusakov, Vorotilov, Wang, Yu, Benhalloum, Mialon, and Scialom}]{froger2025arescalingagentenvironments}
Romain Froger, Pierre Andrews, Matteo Bettini, Amar Budhiraja, Ricardo~Silveira Cabral, Virginie Do, Emilien Garreau, Jean-Baptiste Gaya, Hugo Lauren{\c c}on, Maxime Lecanu, Kunal Malkan, Dheeraj Mekala, Pierre M{\'e}nard, Gerard Moreno-Torres Bertran, Ulyana Piterbarg, Mikhail Plekhanov, Mathieu Rita, Andrey Rusakov, Vladislav Vorotilov, and 5 others. 2025.
\newblock \href {https://arxiv.org/abs/2509.17158} {{ARE: Scaling Up Agent Environments and Evaluations}}.
\newblock \emph{Preprint}, arXiv:2509.17158.

\bibitem[{{Google DeepMind}(2026)}]{google2026gemini38flash}
{Google DeepMind}. 2026.
\newblock \href {https://deepmind.google/models/model-cards/gemini-3-8-flash/} {{Gemini 3.8 Flash}}.
\newblock Accessed September 21, 2026.

\bibitem[{Graves et~al.(2017)Graves, Bellemare, Menick, Munos, and Kavukcuoglu}]{pmlr-v70-graves17a}
Alex Graves, Marc~G. Bellemare, Jacob Menick, R{\'e}mi Munos, and Koray Kavukcuoglu. 2017.
\newblock \href {https://proceedings.mlr.press/v70/graves17a.html} {{Automated Curriculum Learning for Neural Networks}}.
\newblock In \emph{{Proceedings of the 34th International Conference on Machine Learning}}, volume~70 of \emph{Proceedings of Machine Learning Research}, pages 1311--1320. PMLR.

\bibitem[{Guo et~al.(2025)Guo, Yang, Chen, Xiao, Wang, Juan, Qiu, Shen, and Wang}]{guo2025genenvdifficultyalignedcoevolutionllm}
Jiacheng Guo, Ling Yang, Peter Chen, Qixin Xiao, Yinjie Wang, Xinzhe Juan, Jiahao Qiu, Ke~Shen, and Mengdi Wang. 2025.
\newblock \href {https://arxiv.org/abs/2512.19682} {{GenEnv: Difficulty-Aligned Co-Evolution Between LLM Agents and Environment Simulators}}.
\newblock \emph{Preprint}, arXiv:2512.19682.

\bibitem[{He et~al.(2025)He, Sun, Hao, Hao, Xia, Gu, Han, Zhao, Su, Zhang, Gao, Su, Cai, Cai, Yang, and Zhao}]{he2025vitabenchbenchmarkingllmagents}
Wei He, Yueqing Sun, Hongyan Hao, Xueyuan Hao, Zhikang Xia, Qi~Gu, Chengcheng Han, Dengchang Zhao, Hui Su, Kefeng Zhang, Man Gao, Xi~Su, Xiaodong Cai, Xunliang Cai, Yu~Yang, and Yunke Zhao. 2025.
\newblock \href {https://arxiv.org/abs/2509.26490} {{VitaBench: Benchmarking LLM Agents with Versatile Interactive Tasks in Real-world Applications}}.
\newblock \emph{Preprint}, arXiv:2509.26490.

\bibitem[{Hu et~al.(2025)Hu, Zhao, Xu, Sun, Lou, Lin, Luo, and Rajmohan}]{hu2025agentgenenhancingplanningabilities}
Mengkang Hu, Pu~Zhao, Can Xu, Qingfeng Sun, Jianguang Lou, Qingwei Lin, Ping Luo, and Saravan Rajmohan. 2025.
\newblock \href {https://arxiv.org/abs/2408.00764} {{AgentGen: Enhancing Planning Abilities for Large Language Model based Agent via Environment and Task Generation}}.
\newblock \emph{Preprint}, arXiv:2408.00764.

\bibitem[{{Kimi Team} et~al.(2026){Kimi Team}, Bai, Bai, Bao, C., Cai, Cai, Cao, Cao, Chai, Charles, Che, Chen, Chen, Chen, Chen, Chen, Chen, Chen, Chen, Chen, Chen, Chen, Chen, Chen, Chen, Chen, Chen, Chen, Chen, Chen, Chen, Cheng, Cheng, Cui, Cui, Dai, Deng, Ding, Ding, Ding, Dong, Dong, Dong, Dong, Du, Du, Du, Du, Du, Fan, Feng, Feng, Feng, Fu, Fu, Gao, Gao, Gao, Gao, Gao, Geng, Gong, Gong, Gong, Gong, Gu, Gu, Guan, Guo, Guo, Guo, Guo, Hao, Hao, Hao, He, He, He, He, He, He, He, He, He, Hong, Hong, Hu, Hu, Hu, Hu, Hu, Hu, Hua, Huang, Huang, Huang, Huang, Huang, Huang, Huang, Huang, Hui, Jia, Jiang, Jiang, Jiang, Jin, Jin, Jing, Kong, Lai, Li, Li, Li, Li, Li, Li, Li, Li, Li, Li, Li, Li, Li, Li, Li, Li, Li, Li, Li, Li, Li, Li, Li, Li, Lin, Lin, Lin, Lin, Lin, Liu, Liu, Liu, Liu, Liu, Liu, Liu, Liu, Liu, Liu, Liu, Liu, Liu, Liu, Liu, Lu, Lu, Lu, Lu, Lu, Luo, Luo, Luo, Luo, Lyu, Lyu, Mao, Mei, Men, Ni, Niu, Pan, Peng, Qi, Qin, Qin, Qin, Qiu, Qiu, Qiu, Qu, Qu, Shang, Shao, Shen, Shi, Shi, Shi, Shi, Siu, Song, Song, Su, Su, Su, Sui, Sun, Sun, Sun, Sun, Sun, Sun, Tai, Tang, Tang, Tang, Tang, Tian, Tian, Tian, Tu, Wang, Wang, Wang, Wang, Wang, Wang, Wang, Wang, Wang, Wang, Wang, Wang, Wang, Wang, Wang, Wang, Wang, Wang, Wang, Wang, Wang, Wang, Wang, Wang, Wang, Wang, Wang, Wang, Wang, Wang, Wang, Wang, Wang, Wang, Wang, Wang, Wang, Wang, Wang, Wang, Wang, Wei, Wei, Wei, Wen, Wu, Wu, Wu, Wu, Wu, Wu, Wu, Wu, Wu, Xian, Xiang, Xiang, Xiao, Xiao, Xiao, Xie, Xie, Xie, Xie, Xing, Xiong, Xu, Xu, Xu, Xu, Xu, Xu, Xu, Xu, Xu, Xu, Xu, Xu, Xu, Xu, Xu, Xu, Xu, Xu, Xue, Yan, Yan, Yang, Yang, Yang, Yang, Yang, Yang, Yang, Yang, Yang, Yang, Yang, Yang, Yang, Yang, Yang, Yang, Yao, Ye, Ye, Ye, Ye, Yin, Yin, Yin, Yu, Yu, Yu, Yu, Yu, Yu, Yuan, Yuan, Yue, Yue, Yue, Zha, Zhan, Zhang, Zhang, Zhang, Zhang, Zhang, Zhang, Zhang, Zhang, Zhang, Zhang, Zhang, Zhang, Zhang, Zhang, Zhang, Zhang, Zhang, Zhang, Zhang, Zhang, Zhang, Zhang, Zhang, Zhang, Zhang, Zhang, Zhang, Zhang, Zhang, Zhang, Zhang, Zhao, Zhao, Zhao, Zhao, Zhao, Zhao, Zhao, Zhao, Zhao, Zhao, Zhao, Zheng, Zheng, Zheng, Zheng, Zheng, Zhong, Zhong, Zhong, Zhou, Zhou, Zhou, Zhou, Zhou, Zhou, Zhou, Zhu, Zhu, Zhu, Zhu, Zhu, Zhu, Zhuang, Zhuang, and Zu}]{kimiteam2026kimik3openfrontier}
{Kimi Team}, Tongtong Bai, Yifan Bai, Yiping Bao, M.~C., Jianfeng Cai, Xinyuan Cai, Peizhou Cao, Yuxuan Cao, Ziwei Chai, Y.~Charles, H.~S. Che, Guanduo Chen, Guangyu Chen, Guanzheng Chen, Huarong Chen, Jia Chen, Jianlong Chen, Jun Chen, and 383 others. 2026.
\newblock \href {https://arxiv.org/abs/2607.24653} {{Kimi K3: Open Frontier Intelligence}}.
\newblock \emph{Preprint}, arXiv:2607.24653.

\bibitem[{Kumar et~al.(2010)Kumar, Packer, and Koller}]{NIPS2010_e57c6b95}
M.~Pawan Kumar, Benjamin Packer, and Daphne Koller. 2010.
\newblock \href {https://proceedings.neurips.cc/paper_files/paper/2010/file/e57c6b956a6521b28495f2886ca0977a-Paper.pdf} {{Self-Paced Learning for Latent Variable Models}}.
\newblock In \emph{{Advances in Neural Information Processing Systems}}, volume~23. Curran Associates, Inc.

\bibitem[{Li et~al.(2026)Li, Zhao, Zhao, Zeng, Wu, Wang, Ge, Cao, Huang, Liu, Liu, Su, Guo, Zhou, Zhang, Michelini, Wang, Yue, Zhou, Neubig, and He}]{li2026tooldecathlonbenchmarkinglanguage}
Junlong Li, Wenshuo Zhao, Jian Zhao, Weihao Zeng, Haoze Wu, Xiaochen Wang, Rui Ge, Yuxuan Cao, Yuzhen Huang, Wei Liu, Junteng Liu, Zhaochen Su, Yiyang Guo, Fan Zhou, Lueyang Zhang, Juan Michelini, Xingyao Wang, Xiang Yue, Shuyan Zhou, and 2 others. 2026.
\newblock \href {https://arxiv.org/abs/2510.25726} {{The Tool Decathlon: Benchmarking Language Agents for Diverse, Realistic, and Long-Horizon Task Execution}}.
\newblock \emph{Preprint}, arXiv:2510.25726.

\bibitem[{Li et~al.(2025)Li, Inan, Yue, Chen, Wutschitz, Kulkarni, Poovendran, Sim, and Rajmohan}]{li2025simulatingenvironmentsreasoningmodels}
Yuetai Li, Huseyin~A Inan, Xiang Yue, Wei-Ning Chen, Lukas Wutschitz, Janardhan Kulkarni, Radha Poovendran, Robert Sim, and Saravan Rajmohan. 2025.
\newblock \href {https://arxiv.org/abs/2511.01824} {{Simulating Environments with Reasoning Models for Agent Training}}.
\newblock \emph{Preprint}, arXiv:2511.01824.

\bibitem[{Liu et~al.(2025)Liu, Yu, Zhang, Xu, Lei, Lai, Gu, Ding, Men, Yang, Zhang, Deng, Zeng, Du, Zhang, Shen, Zhang, Su, Sun, Huang, Dong, and Tang}]{liu2025agentbenchevaluatingllmsagents}
Xiao Liu, Hao Yu, Hanchen Zhang, Yifan Xu, Xuanyu Lei, Hanyu Lai, Yu~Gu, Hangliang Ding, Kaiwen Men, Kejuan Yang, Shudan Zhang, Xiang Deng, Aohan Zeng, Zhengxiao Du, Chenhui Zhang, Sheng Shen, Tianjun Zhang, Yu~Su, Huan Sun, and 3 others. 2025.
\newblock \href {https://arxiv.org/abs/2308.03688} {{AgentBench: Evaluating LLMs as Agents}}.
\newblock \emph{Preprint}, arXiv:2308.03688.

\bibitem[{{OpenAI}(n.d.)}]{openai_gpt56luna}
{OpenAI}. n.d.
\newblock \href {https://developers.openai.com/api/docs/models/gpt-5.6-luna} {{GPT-5.6 Luna}}.
\newblock Accessed September 21, 2026.

\bibitem[{Qi et~al.(2025)Qi, Liu, Iong, Lai, Sun, Zhao, Yang, Yang, Sun, Yao, Zhang, Xu, Tang, and Dong}]{qi2025webrltrainingllmweb}
Zehan Qi, Xiao Liu, Iat~Long Iong, Hanyu Lai, Xueqiao Sun, Wenyi Zhao, Yu~Yang, Xinyue Yang, Jiadai Sun, Shuntian Yao, Tianjie Zhang, Wei Xu, Jie Tang, and Yuxiao Dong. 2025.
\newblock \href {https://arxiv.org/abs/2411.02337} {{WebRL: Training LLM Web Agents via Self-Evolving Online Curriculum Reinforcement Learning}}.
\newblock \emph{Preprint}, arXiv:2411.02337.

\bibitem[{Qin et~al.(2023)Qin, Liang, Ye, Zhu, Yan, Lu, Lin, Cong, Tang, Qian, Zhao, Hong, Tian, Xie, Zhou, Gerstein, Li, Liu, and Sun}]{qin2023toolllmfacilitatinglargelanguage}
Yujia Qin, Shihao Liang, Yining Ye, Kunlun Zhu, Lan Yan, Yaxi Lu, Yankai Lin, Xin Cong, Xiangru Tang, Bill Qian, Sihan Zhao, Lauren Hong, Runchu Tian, Ruobing Xie, Jie Zhou, Mark Gerstein, Dahai Li, Zhiyuan Liu, and Maosong Sun. 2023.
\newblock \href {https://arxiv.org/abs/2307.16789} {{ToolLLM: Facilitating Large Language Models to Master 16000+ Real-world APIs}}.
\newblock \emph{Preprint}, arXiv:2307.16789.

\bibitem[{{Qwen Team}(n.d.)}]{qwen38modelcard}
{Qwen Team}. n.d.
\newblock \href {https://huggingface.co/Qwen/Qwen3.8-2.4T-A95B/blob/main/README.md} {{Qwen3.8-2.4T-A95B}}.
\newblock Accessed September 21, 2026.

\bibitem[{Schick et~al.(2023)Schick, Dwivedi-Yu, Dessi, Raileanu, Lomeli, Hambro, Zettlemoyer, Cancedda, and Scialom}]{schick2023toolformerlanguagemodelsteach}
Timo Schick, Jane Dwivedi-Yu, Roberto Dessi, Roberta Raileanu, Maria Lomeli, Eric Hambro, Luke Zettlemoyer, Nicola Cancedda, and Thomas Scialom. 2023.
\newblock \href {https://doi.org/10.52202/075280-2997} {{Toolformer: Language Models Can Teach Themselves to Use Tools}}.
\newblock In \emph{{Advances in Neural Information Processing Systems}}, volume~36, pages 68539--68551. Curran Associates, Inc.

\bibitem[{Shi et~al.(2026)Shi, Zytek, Razavi, Narasimhan, and Barres}]{shi2026tauknowledge}
Quan Shi, Alexandra Zytek, Pedram Razavi, Karthik Narasimhan, and Victor Barres. 2026.
\newblock \href {https://arxiv.org/abs/2603.04370} {{\ensuremath{\tau}-Knowledge: Evaluating Conversational Agents over Unstructured Knowledge}}.
\newblock \emph{Preprint}, arXiv:2603.04370.

\bibitem[{Shinn et~al.(2023)Shinn, Cassano, Gopinath, Narasimhan, and Yao}]{shinn2023reflexion}
Noah Shinn, Federico Cassano, Ashwin Gopinath, Karthik Narasimhan, and Shunyu Yao. 2023.
\newblock \href {https://doi.org/10.52202/075280-0377} {{Reflexion: language agents with verbal reinforcement learning}}.
\newblock In \emph{{Advances in Neural Information Processing Systems}}, volume~36, pages 8634--8652. Curran Associates, Inc.

\bibitem[{Song et~al.(2026)Song, Chang, Dong, Zhu, Wen, and Dou}]{song2026envscalerscalingtoolinteractiveenvironments}
Xiaoshuai Song, Haofei Chang, Guanting Dong, Yutao Zhu, Ji-Rong Wen, and Zhicheng Dou. 2026.
\newblock \href {https://arxiv.org/abs/2601.05808} {{EnvScaler: Scaling Tool-Interactive Environments for LLM Agent via Programmatic Synthesis}}.
\newblock \emph{Preprint}, arXiv:2601.05808.

\bibitem[{Trivedi et~al.(2024)Trivedi, Khot, Hartmann, Manku, Dong, Li, Gupta, Sabharwal, and Balasubramanian}]{trivedi2024appworldcontrollableworldapps}
Harsh Trivedi, Tushar Khot, Mareike Hartmann, Ruskin Manku, Vinty Dong, Edward Li, Shashank Gupta, Ashish Sabharwal, and Niranjan Balasubramanian. 2024.
\newblock \href {https://arxiv.org/abs/2407.18901} {{AppWorld: A Controllable World of Apps and People for Benchmarking Interactive Coding Agents}}.
\newblock \emph{Preprint}, arXiv:2407.18901.

\bibitem[{Tu et~al.(2026)Tu, Hao, Yang, Chen, Zhang, Xia, Yang, Sun, Liu, Shen, Gu, Su, and Cai}]{tu2026scaleenvscalingenvironmentsynthesis}
Dunwei Tu, Hongyan Hao, Hansi Yang, Yihao Chen, Yi-Kai Zhang, Zhikang Xia, Yu~Yang, Yueqing Sun, Xingchen Liu, Furao Shen, Qi~Gu, Hui Su, and Xunliang Cai. 2026.
\newblock \href {https://arxiv.org/abs/2602.06820} {{ScaleEnv: Scaling Environment Synthesis from Scratch for Generalist Interactive Tool-Use Agent Training}}.
\newblock \emph{Preprint}, arXiv:2602.06820.

\bibitem[{Wang et~al.(2026)Wang, Xu, Liu, Wang, Han, Yao, Yao, and He}]{wang2026agentworldmodelinfinity}
Zhaoyang Wang, Canwen Xu, Boyi Liu, Yite Wang, Siwei Han, Zhewei Yao, Huaxiu Yao, and Yuxiong He. 2026.
\newblock \href {https://arxiv.org/abs/2602.10090} {{Agent World Model: Infinity Synthetic Environments for Agentic Reinforcement Learning}}.
\newblock \emph{Preprint}, arXiv:2602.10090.

\bibitem[{Wang et~al.(2024)Wang, Mao, Fried, and Neubig}]{wang2024agentworkflowmemory}
Zora~Zhiruo Wang, Jiayuan Mao, Daniel Fried, and Graham Neubig. 2024.
\newblock \href {https://arxiv.org/abs/2409.07429} {{Agent Workflow Memory}}.
\newblock \emph{Preprint}, arXiv:2409.07429.

\bibitem[{Wu et~al.(2025)Wu, Liu, Zhang, Chen, Meng, Du, Zhao, Zhang, Ye, Wang, Wang, Ni, Yang, Xu, and Shieh}]{wu2025mcpmarkbenchmarkstresstestingrealistic}
Zijian Wu, Xiangyan Liu, Xinyuan Zhang, Lingjun Chen, Fanqing Meng, Lingxiao Du, Yiran Zhao, Fanshi Zhang, Yaoqi Ye, Jiawei Wang, Zirui Wang, Jinjie Ni, Yufan Yang, Arvin Xu, and Michael~Qizhe Shieh. 2025.
\newblock \href {https://arxiv.org/abs/2509.24002} {{MCPMark: A Benchmark for Stress-Testing Realistic and Comprehensive MCP Use}}.
\newblock \emph{Preprint}, arXiv:2509.24002.

\bibitem[{Xi et~al.(2025)Xi, Huang, Liao, Huang, Guo, Liu, Zheng, Ye, Zhang, Chen, He, Ding, Li, Chen, Du, Yao, Xu, Chen, Gui, Wu, Zhang, Huang, and Jiang}]{xi2025agentgymrltrainingllmagents}
Zhiheng Xi, Jixuan Huang, Chenyang Liao, Baodai Huang, Honglin Guo, Jiaqi Liu, Rui Zheng, Junjie Ye, Jiazheng Zhang, Wenxiang Chen, Wei He, Yiwen Ding, Guanyu Li, Zehui Chen, Zhengyin Du, Xuesong Yao, Yufei Xu, Jiecao Chen, Tao Gui, and 4 others. 2025.
\newblock \href {https://arxiv.org/abs/2509.08755} {{AgentGym-RL: Training LLM Agents for Long-Horizon Decision Making through Multi-Turn Reinforcement Learning}}.
\newblock \emph{Preprint}, arXiv:2509.08755.

\bibitem[{Xie et~al.(2024)Xie, Zhang, Chen, Li, Zhao, Cao, Hua, Cheng, Shin, Lei, Liu, Xu, Zhou, Savarese, Xiong, Zhong, and Yu}]{xie2024osworld}
Tianbao Xie, Danyang Zhang, Jixuan Chen, Xiaochuan Li, Siheng Zhao, Ruisheng Cao, Toh~Jing Hua, Zhoujun Cheng, Dongchan Shin, Fangyu Lei, Yitao Liu, Yiheng Xu, Shuyan Zhou, Silvio Savarese, Caiming Xiong, Victor Zhong, and Tao Yu. 2024.
\newblock \href {https://doi.org/10.52202/079017-1650} {{OSWorld: Benchmarking Multimodal Agents for Open-Ended Tasks in Real Computer Environments}}.
\newblock In \emph{{Advances in Neural Information Processing Systems}}, volume~37, pages 52040--52094. Curran Associates, Inc.

\bibitem[{Xu et~al.(2026{\natexlab{a}})Xu, Hu, Zhang, Zhang, Zhang, Demoret, Kautz, and Dong}]{xu2026reinforcingcollectiveskills}
Binfeng Xu, Jian Hu, Hao Zhang, Shaokun Zhang, Yifan Zhang, Michael Demoret, Jan Kautz, and Yi~Dong. 2026{\natexlab{a}}.
\newblock \href {https://www.alphaxiv.org/abs/2609.reinforcing-agents-collective-skills} {{Reinforcing Agents with Collective Skills}}.
\newblock alphaXiv preprint.

\bibitem[{Xu et~al.(2026{\natexlab{b}})Xu, Wang, DENG, Li, Yang, Zhu, Liu, Zhu, Huang, Chen, Deng, Mi, Shang, Zeng, and Guo}]{xu2026envfactoryscalingtooluseagents}
Minrui Xu, Zilin Wang, Mengyi DENG, Zhiwei Li, Zhicheng Yang, Xiao Zhu, Yinhong Liu, Boyu Zhu, Baiyu Huang, Chao Chen, Heyuan Deng, Fei Mi, Lifeng Shang, Xingshan Zeng, and Zhijiang Guo. 2026{\natexlab{b}}.
\newblock \href {https://arxiv.org/abs/2605.18703} {{EnvFactory: Scaling Tool-Use Agents via Executable Environments Synthesis and Robust RL}}.
\newblock \emph{Preprint}, arXiv:2605.18703.

\bibitem[{Xu et~al.(2025{\natexlab{a}})Xu, Chen, Wu, Chen, Zhang, Yao, Xie, Chen, Liu, Qian, Yang, Jin, Deng, Torr, Ghanem, and Li}]{xu-etal-2025-crab}
Tianqi Xu, Linyao Chen, Dai-Jie Wu, Yanjun Chen, Zecheng Zhang, Xiang Yao, Zhiqiang Xie, Yongchao Chen, Shilong Liu, Bochen Qian, Anjie Yang, Zhaoxuan Jin, Jianbo Deng, Philip Torr, Bernard Ghanem, and Guohao Li. 2025{\natexlab{a}}.
\newblock \href {https://doi.org/10.18653/v1/2025.findings-acl.1113} {{CRAB}: Cross-environment agent benchmark for multimodal language model agents}.
\newblock In \emph{Findings of the Association for Computational Linguistics: ACL 2025}, pages 21607--21647, Vienna, Austria. Association for Computational Linguistics.

\bibitem[{Xu et~al.(2025{\natexlab{b}})Xu, Lu, Shen, Wang, Wang, Mao, Xiong, and Yu}]{xu2025agenttrekagenttrajectorysynthesis}
Yiheng Xu, Dunjie Lu, Zhennan Shen, Junli Wang, Zekun Wang, Yuchen Mao, Caiming Xiong, and Tao Yu. 2025{\natexlab{b}}.
\newblock \href {https://arxiv.org/abs/2412.09605} {{AgentTrek: Agent Trajectory Synthesis via Guiding Replay with Web Tutorials}}.
\newblock \emph{Preprint}, arXiv:2412.09605.

\bibitem[{Xu et~al.(2025{\natexlab{c}})Xu, Soria, Tan, Roy, Agrawal, Poovendran, and Panda}]{xu2025toucansynthesizing15mtoolagentic}
Zhangchen Xu, Adriana~Meza Soria, Shawn Tan, Anurag Roy, Ashish~Sunil Agrawal, Radha Poovendran, and Rameswar Panda. 2025{\natexlab{c}}.
\newblock \href {https://arxiv.org/abs/2510.01179} {{TOUCAN: Synthesizing 1.5M Tool-Agentic Data from Real-World MCP Environments}}.
\newblock \emph{Preprint}, arXiv:2510.01179.

\bibitem[{Xu et~al.(2026{\natexlab{c}})Xu, Yao, Chen, Guo, Lu, Lu, Huang, Xu, Wang, Zhang, Zeng, and Liu}]{xu2026envaceinternalizingenvironmentdynamics}
Zishan Xu, Zhiyuan Yao, Yuxin Chen, Yifu Guo, Zhengxi Lu, Yuquan Lu, Jinyang Huang, Yan Xu, Yasheng Wang, Weinan Zhang, Xingshan Zeng, and Weiwen Liu. 2026{\natexlab{c}}.
\newblock \href {https://arxiv.org/abs/2608.06197} {{EnvACE: Internalizing Environment Dynamics via World Rehearsal for Agentic Reinforcement Learning}}.
\newblock \emph{Preprint}, arXiv:2608.06197.

\bibitem[{Yang et~al.(2025{\natexlab{a}})Yang, Li, Yang, Zhang, Hui, Zheng, Yu, Gao, Huang, Lv, Zheng, Liu, Zhou, Huang, Hu, Ge, Wei, Lin, Tang, Yang, Tu, Zhang, Yang, Yang, Zhou, Zhou, Lin, Dang, Bao, Yang, Yu, Deng, Li, Xue, Li, Zhang, Wang, Zhu, Men, Gao, Liu, Luo, Li, Tang, Yin, Ren, Wang, Zhang, Ren, Fan, Su, Zhang, Zhang, Wan, Liu, Wang, Cui, Zhang, Zhou, and Qiu}]{yang2025qwen3technicalreport}
An~Yang, Anfeng Li, Baosong Yang, Beichen Zhang, Binyuan Hui, Bo~Zheng, Bowen Yu, Chang Gao, Chengen Huang, Chenxu Lv, Chujie Zheng, Dayiheng Liu, Fan Zhou, Fei Huang, Feng Hu, Hao Ge, Haoran Wei, Huan Lin, Jialong Tang, and 41 others. 2025{\natexlab{a}}.
\newblock \href {https://arxiv.org/abs/2505.09388} {{Qwen3 Technical Report}}.
\newblock \emph{Preprint}, arXiv:2505.09388.

\bibitem[{Yang et~al.(2025{\natexlab{b}})Yang, Le, Xing, An, Chen, Zhao, Song, and Zhang}]{yang2025toolmindtechnicalreportlargescale}
Chen Yang, Ran Le, Yun Xing, Zhenwei An, Zongchao Chen, Wayne~Xin Zhao, Yang Song, and Tao Zhang. 2025{\natexlab{b}}.
\newblock \href {https://arxiv.org/abs/2511.15718} {{ToolMind Technical Report: A Large-Scale, Reasoning-Enhanced Tool-Use Dataset}}.
\newblock \emph{Preprint}, arXiv:2511.15718.

\bibitem[{Yang et~al.(2025{\natexlab{c}})Yang, Lieret, Jimenez, Wettig, Khandpur, Zhang, Hui, Press, Schmidt, and Yang}]{yang2025swesmithscalingdatasoftware}
John Yang, Kilian Lieret, Carlos~E. Jimenez, Alexander Wettig, Kabir Khandpur, Yanzhe Zhang, Binyuan Hui, Ofir Press, Ludwig Schmidt, and Diyi Yang. 2025{\natexlab{c}}.
\newblock \href {https://doi.org/10.52202/085713-3239} {{SWE-smith: Scaling Data for Software Engineering Agents}}.
\newblock In \emph{{Advances in Neural Information Processing Systems}}, volume 38, Main Conference. Curran Associates, Inc.

\bibitem[{Yao et~al.(2024)Yao, Shinn, Razavi, and Narasimhan}]{yao2024taubenchbenchmarktoolagentuserinteraction}
Shunyu Yao, Noah Shinn, Pedram Razavi, and Karthik Narasimhan. 2024.
\newblock \href {https://arxiv.org/abs/2406.12045} {{$\tau$-bench: A Benchmark for Tool-Agent-User Interaction in Real-World Domains}}.
\newblock \emph{Preprint}, arXiv:2406.12045.

\bibitem[{Yao et~al.(2023)Yao, Zhao, Yu, Du, Shafran, Narasimhan, and Cao}]{yao2023reactsynergizingreasoningacting}
Shunyu Yao, Jeffrey Zhao, Dian Yu, Nan Du, Izhak Shafran, Karthik Narasimhan, and Yuan Cao. 2023.
\newblock \href {https://arxiv.org/abs/2210.03629} {{ReAct: Synergizing Reasoning and Acting in Language Models}}.
\newblock \emph{Preprint}, arXiv:2210.03629.

\bibitem[{Ye et~al.(2026)Ye, Li, Yang, Liu, Yao, Lv, Xie, An, Li, Kong, Liu, Sui, and Yang}]{ye2026clawevaltrustworthyevaluationautonomous}
Bowen Ye, Rang Li, Qibin Yang, Yuanxin Liu, Linli Yao, Hanglong Lv, Zhihui Xie, Chenxin An, Lei Li, Lingpeng Kong, Qi~Liu, Zhifang Sui, and Tong Yang. 2026.
\newblock \href {https://arxiv.org/abs/2604.06132} {Claw-eval: Towards trustworthy evaluation of autonomous agents}.
\newblock \emph{Preprint}, arXiv:2604.06132.

\bibitem[{{Z.ai}(n.d.)}]{zai_glm53}
{Z.ai}. n.d.
\newblock \href {https://huggingface.co/zai-org/GLM-5.3/blob/main/README.md} {{GLM-5.3}}.
\newblock Accessed September 21, 2026.

\bibitem[{Zhang et~al.(2026)Zhang, Jiang, Li, Tu, Su, Deng, Guo, Lv, and Lin}]{zhang2026deepplanningbenchmarkinglonghorizonagentic}
Yinger Zhang, Shutong Jiang, Renhao Li, Jianhong Tu, Yang Su, Lianghao Deng, Xudong Guo, Chenxu Lv, and Junyang Lin. 2026.
\newblock \href {https://arxiv.org/abs/2601.18137} {{DeepPlanning: Benchmarking Long-Horizon Agentic Planning with Verifiable Constraints}}.
\newblock \emph{Preprint}, arXiv:2601.18137.

\bibitem[{Zhou et~al.(2026)Zhou, Yue, Hu, Shen, Zhang, feng hong, Liu, and Zeng}]{zhou2026toolverseunlockingmassiveenvironments}
Shuaiyu Zhou, Fengpeng Yue, Zengjie Hu, Yuanzhe Shen, Chenyang Zhang, feng hong, Cao Liu, and Ke~Zeng. 2026.
\newblock \href {https://arxiv.org/abs/2607.15660} {{ToolVerse: Unlocking Massive Environments and Long-Horizon Tasks for Agentic Reinforcement Learning}}.
\newblock \emph{Preprint}, arXiv:2607.15660.

\bibitem[{Zhou et~al.(2024)Zhou, Xu, Zhu, Zhou, Lo, Sridhar, Cheng, Ou, Bisk, Fried, Alon, and Neubig}]{zhou2024webarenarealisticwebenvironment}
Shuyan Zhou, Frank~F Xu, Hao Zhu, Xuhui Zhou, Robert Lo, Abishek Sridhar, Xianyi Cheng, Tianyue Ou, Yonatan Bisk, Daniel Fried, Uri Alon, and Graham Neubig. 2024.
\newblock \href {https://proceedings.iclr.cc/paper_files/paper/2024/file/4410c0711e9154a7a2d26f9b3816d1ef-Paper-Conference.pdf} {{WebArena: A Realistic Web Environment for Building Autonomous Agents}}.
\newblock In \emph{{International Conference on Learning Representations}}, volume 2024, pages 15585--15606.

\end{thebibliography}
\clearpage
\appendix
\section{Detailed CrossEnv Method}
\label{app:crossenv-details}

Building on the task definition in Equation~\ref{eq:crossenv-task}, this appendix supplements Section~\ref{sec:crossenv-method} with formal interaction and graph definitions, seed sources and taxonomy statistics, tool testing procedures, and detailed data alignment and verification. It also specifies the reference replay and independent-agent filtering criteria used to retain tasks.

\subsection{Agentic Interaction across Environments}
\label{app:environment-interaction}

We consider an agent that completes tasks by invoking tools across multiple environments and interacting with a user when required. Each environment $E_i=(\mathcal S_i,\mathcal T_i,\mathcal O_i,F_i)$ consists of a state space $\mathcal S_i$, a tool set $\mathcal T_i$, an observation space $\mathcal O_i$, and an execution function $F_i$. The state represents the current application data. Let $\mathcal E=\{E_1,\ldots,E_n\}$ denote the available environments. For a task $\tau$, only its participating environments $\mathcal E_S\subseteq\mathcal E$ enter the joint state $\mathbf s_t=(s_t^{(i)})_{E_i\in\mathcal E_S}$, with $\mathbf s_0=\mathbf s_{0,\tau}$. We suppress the task index on subsequent states for readability. At each step, the agent selects an action using the task instruction $I$ and interaction history $h_t$. For a tool call, $a_t=(i_t,f_t,x_t)$, where $E_{i_t}\in\mathcal E_S$, $f_t\in\mathcal T_{i_t}$, and $x_t$ denotes the tool arguments. The selected environment executes the call and returns an observation:
\begin{equation}
\label{eq:crossenv-interaction}
\begin{aligned}
(s_{t+1}^{(i_t)},o_t)&=F_{i_t}(s_t^{(i_t)},(f_t,x_t)),\\
s_{t+1}^{(j)}&=s_t^{(j)}\quad(E_j\in\mathcal E_S,\ j\ne i_t).
\end{aligned}
\end{equation}
For tasks requiring user interaction, $a_t$ may instead be a user-directed message. The task's user interaction interface supplies a reply observation $o_t$ and determines any associated application-state changes. User exchanges are recorded in $h_t$ alongside tool calls and their outputs. Equation~\ref{eq:crossenv-interaction} describes tool execution only.

The agent selects actions from the task instruction and interaction feedback, using earlier tool outputs to coordinate subsequent operations~\citep{yao2023reactsynergizingreasoningacting,wang2024agentworkflowmemory}. Let $A=(a_0,\ldots,a_{T-1})$ denote the action sequence, including tool calls and user-directed interactions where applicable, and let $T=|A|$. The resulting observations $O_A=(o_0,\ldots,o_{T-1})$ include tool outputs and any user replies. The same convention applies to the reference sequence $A^*$. Task completion is evaluated using these observations and the final joint state $\mathbf s_T$ of the participating environments~\citep{trivedi2024appworldcontrollableworldapps,yao2024taubenchbenchmarktoolagentuserinteraction,wu2025mcpmarkbenchmarkstresstestingrealistic}. Each executable rule $c\in C_\tau$ returns a binary result; task completion requires
\begin{equation}
\label{eq:crossenv-completion}
c(O_A,\mathbf s_T)=1,\qquad \forall c\in C_\tau.
\end{equation}

\subsection{Phase 1: Single-Environment Research}
\label{app:environment-research}

\noindent\textbf{Environment theme collection.} Diverse environment themes provide concrete application scenarios and operational context for environment synthesis. We collect environment seeds from MCP services and publicly available agent skill documents. MCP services provide functionality descriptions and tool interfaces, while skill documents describe task-specific procedures and workflows. After filtering the seeds and organizing their capabilities, we obtain \textbf{801 MCP seeds and 682 skill seeds}, covering \textbf{14,657 tool capability descriptions}. These serve as the starting points for environment research and construction.\footnote{MCP seeds are collected from Smithery. Skill sources include Smithery Skills, SkillMD, and the GitHub repositories \nolinkurl{sickn33/agentic-awesome-skills} and \nolinkurl{github/awesome-copilot}.}

\noindent\textbf{Environment taxonomy construction.} To support coherent cross-environment task composition, we organize environment seeds using a two-level taxonomy of \textbf{10 business domains and 84 capability categories}, drawing on the Agent-World taxonomy~\citep{dong2026agentworldscalingrealworldenvironment} and the business characteristics and capabilities of our seeds. Each seed is assigned one primary domain and capability category, while a capability may be associated with multiple domains.

\noindent\textbf{Environment research and data construction.} Given an environment seed, a research agent retrieves relevant documentation and available data using the environment description and tool functionality descriptions. A data construction agent then organizes the environment data from these findings. When information is missing, it sends a request to the research agent for further retrieval. Once sufficient information is available, this process produces the initial database, which represents the initial state $s_0^{(i)}$ of environment $E_i$.

\subsection{Phase 2: Local Dependency Graphs}
\label{app:local-dependency-graph}

\noindent\textbf{Entity extraction.} Using the initial database and tool descriptions, we abstract the application objects queried or manipulated by tools into entity types and organize their descriptive information as attributes. All environments share an entity type set $\mathcal C$, where each type is defined as
\begin{equation}
c=(n_c,\mathcal A_c)\in\mathcal C,\qquad
\mathcal A_c=\bigcup_{i:\,c\in V_i}\mathcal A_c^{(i)},
\end{equation}
where $n_c$ is the entity type name, $\mathcal A_c^{(i)}$ is its attribute set in environment $E_i$, and $\mathcal A_c$ is the union of these attributes across environments. The entity types extracted from $E_i$ form the node set $V_i\subseteq\mathcal C$ of its local dependency graph. This representation connects corresponding entity types across environments without selecting or equating concrete database records; task-specific records are aligned in Phase~5.

\noindent\textbf{Relation extraction.} We then represent tool functionality as relations between entity types and associate each relation with the tools that implement it. In Figure~\ref{fig:pipeline}, for example, \texttt{reserve\_room} supports a booking relation from \texttt{Person} to \texttt{Hotel}, while \texttt{search\_hotels} supports a search relation from \texttt{City} to \texttt{Hotel}. The figure labels these relations with their supporting tool names. This yields a local dependency graph
\begin{equation}
\label{eq:local-entity-tool-graph}
G_i=(V_i,R_i),\qquad r=(u,\rho,v,\mathcal T_r)\in R_i,
\end{equation}
where $V_i$ is the entity type set and $R_i$ is the set of directed relations. Each relation $r$ contains a source entity type $u\in V_i$, a relation label $\rho$ describing its semantics, a target entity type $v\in V_i$, and a set of supporting tools $\mathcal T_r\subseteq\mathcal T_i$.

\subsection{Phase 3: Cross-Environment Graph Composition}
\label{app:graph-composition}

\noindent\textbf{Connecting environments through shared entity types.} Tools in different environments may provide different operations on the same entity type. We merge the local graphs through shared entity types while retaining the environment associated with each relation and its tools. The resulting cross-environment dependency graph is
\begin{equation}
\label{eq:crossenv-graph-composition}
\begin{aligned}
G^\times&=(V^\times,R^\times),\\
V^\times&=\bigcup_i V_i,\\
R^\times&=\bigcup_i\{(i,r)\mid r\in R_i\}.
\end{aligned}
\end{equation}
Each shared entity type corresponds to a single node in the composed graph. Its attributes aggregate information from different environments, while its incident edges retain the operations supported by each environment. To incorporate a new environment, we extract its entities and relations and connect its local graph to the existing graph through shared entity types. We can then sample task combinations that include the new environment.

\noindent\textbf{Task skeleton sampling.} We select connected relations and their associated tools from the cross-environment graph to form task skeletons. Candidate combinations are assessed along three dimensions: \textbf{entity relationships}, which indicate whether operations in different environments can involve common objects; \textbf{complementary operations}, which indicate whether create, read, update, and delete (CRUD) operations can form a coherent workflow; and \textbf{business relevance}, which indicates whether the capabilities naturally belong in the same task. Filtering candidates along these dimensions yields a task skeleton
\begin{equation}
\begin{gathered}
S=(V_S,R_S),\\
V_S\subseteq V^\times,\qquad R_S\subseteq R^\times,
\end{gathered}
\end{equation}
where $(V_S,R_S)$ is a connected subgraph specifying the entity types, relations, and supporting tools. Its participating environments are $\mathcal E_S=\{E_i\mid(i,r)\in R_S\}$, and $\mathcal T_S$ denotes their selected supporting tools. The index $i$ in each relation identifies its source environment. We then assign a task objective to the selected combination, specifying what the operations should accomplish together. This determines which operations must use earlier results and which cross-environment constraints the final outcomes must satisfy. The subgraph specifies the skeleton's structure, while the objective and constraints supply its task semantics. If the workflow requires additional operations, we record the corresponding tool requirements.

\noindent\textbf{Structural difficulty control.} We control task skeleton complexity along two dimensions: the number of environments and their semantic scope. Restricting the number of environments controls the scale of coordination required by a task. Restricting their business domains controls its semantic scope, from coordination within a single domain to coordination across domains.

\noindent\textbf{User interaction.} We additionally specify whether a task requires interaction with the user. In CrossEnv-Curriculum, L5 introduces user interaction while retaining L4's environment-count and business-domain-scope requirements, as described in Section~\ref{sec:crossenv-curriculum}.

\subsection{Phase 4: Tool Synthesis and Testing}
\label{app:tool-synthesis-testing}

\noindent\textbf{Tool synthesis.} Given a task skeleton, we group the required tools by environment, reuse existing implementations that have passed testing, and use a coding agent to implement missing tools. The coding agent uses tool descriptions, research findings, and the initial database to generate executable code that reads or modifies data and returns results. This turns the tool capabilities associated with graph relations into executable operations. For environment $E_i$, the tested implementations define the execution function $F_i$ introduced in Section~\ref{app:environment-interaction}.

\noindent\textbf{Tool testing.} We invoke the generated tools and check whether their outputs and data changes match the expected behavior. For read operations, we check that the returned information is correct and the data remains unchanged. For create, update, and delete operations, we check that the intended changes occur and confirm them through subsequent queries. We also test invalid inputs and conflicting operations to assess error handling. Failures are returned to the coding agent for correction, followed by another round of testing. The executable environments are considered complete only when all tools required by the skeleton pass these tests.

\subsection{Phase 5: Task Instantiation and Verification}
\label{app:task-instantiation-verification}

\noindent\textbf{Task instantiation.} Given a task skeleton $S$, its objective, and its constraints, we select concrete entities and data for the task. Because the initial states $s_0^{(i)}$ are constructed independently, records for the same object may be missing or contain conflicting information across relevant environments. We supplement required records and resolve conflicts so that the selected objects and cross-environment constraints are consistent. Initial data and reference actions are planned jointly: records needed by read, update, and delete operations must either exist initially or be created by earlier actions; records that the task requires the agent to create are not inserted in advance. This produces a task-specific initial state $s_{0,\tau}^{(i)}$ for each environment and the joint initial state
\begin{equation}
\mathbf s_{0,\tau}=\bigl(s_{0,\tau}^{(i)}\bigr)_{E_i\in\mathcal E_S}.
\end{equation}
\noindent\textbf{Reference execution and task generation.} Starting from $\mathbf s_{0,\tau}$, we execute the planned actions, including user interaction where required, and record their observations and data changes. Later actions can use results produced by earlier operations. If execution diverges from expectations, we revise the task data and action plan based on the feedback and execute the plan again. Once execution succeeds, we retain the reference action sequence $A^*$ and generate the task instruction $I$ from the scenario and execution results.

\noindent\textbf{Verification rule generation.} We derive executable verification rules $C_\tau$ from the task objective and constraints, using the tool outputs and data changes observed during reference execution to specify expected outcomes. These rules check whether the required records have been created, updated, or deleted; whether returned results satisfy the specified conditions; and whether related information is consistent across environments. We then construct negative cases, such as omitted operations and incorrect values, to check whether the rules reject incomplete or incorrect task outcomes.

\noindent\textbf{Task verification.} For an action sequence $A$, let $T(A)=|A|$ and define its binary verification outcome as
\begin{equation}
\operatorname{Pass}_\tau(A)=\prod_{c\in C_\tau}c(O_A,\mathbf s_{T(A)}),
\end{equation}
where execution starts from $\mathbf s_{0,\tau}$ and $\mathbf s_{T(A)}$ contains the final states of $\mathcal E_S$. This is one exactly when every verification rule passes. We use two levels of verification: reference action replay and independent agent attempts. First, we reset all participating environments to $\mathbf s_{0,\tau}$, replay $A^*$, and require reproducible results with $\operatorname{Pass}_\tau(A^*)=1$. We then run five independent agent attempts using the task instruction, available tools, and user interaction where required, with every attempt starting from the same initial state. Let $A^{(k)}$ denote the action sequence from attempt $k$. Following successful reference replay, we retain the task if $\sum_{k=1}^{5}\mathbf 1[\operatorname{Pass}_\tau(A^{(k)})]\ge 2$. These two checks assess the reproducibility of the reference solution and whether an agent can find a valid solution from the task instruction and interfaces, respectively.

The retained task follows Equation~\ref{eq:crossenv-task}, with aligned initial states $\mathbf s_{0,\tau}$, a successful reference sequence $A^*$, and executable verification rules $C_\tau$.
\section{Environment Taxonomy}
\label{app:environment-taxonomy}

Table~\ref{tab:environment-taxonomy} lists 10 business domains ($C_1$) and 84 capability categories ($C_2$). Categories may span domains; each seed has one primary domain and category, assigned by its main purpose.

\begin{table*}[t]
\centering
\small
\setlength{\tabcolsep}{5pt}
\renewcommand{\arraystretch}{1.15}
\begin{tabularx}{\linewidth}{@{}>{\raggedright\arraybackslash}p{0.20\linewidth}>{\raggedright\arraybackslash}X@{}}
\toprule
\textbf{Business domain ($C_1$)} & \textbf{Capability categories ($C_2$)} \\
\midrule
\textbf{Travel} & Flight booking; Accommodation booking; Food and dining; Transit and navigation; Earth science and environmental monitoring; Maps and location services; Weather and forecasting; Itinerary planning; Currency exchange; Messaging and notification; Task and project management; Local search; Calendar and scheduling. \\
\addlinespace[4pt]
\textbf{Work \& Office} & Decision support; Messaging and notification; Task and project management; Calendar and scheduling; Email management; Document processing; Presentation and design; Knowledge management; Web search; Customer support. \\
\addlinespace[4pt]
\textbf{Life Services} & Food and dining; Transit and navigation; Earth science and environmental monitoring; Maps and location services; Weather and forecasting; Messaging and notification; Calendar and scheduling; Shopping; Home services; Local news; Medical appointment booking; Billing and payments; Arts, culture, and museums; Events and entertainment; Gaming and interactive experiences; Real estate services. \\
\addlinespace[4pt]
\textbf{Finance \& Investment} & Currency exchange; Billing and payments; Crypto market data; Market intelligence; Stock and equity investment; Blockchain infrastructure and wallets; DeFi and DEX protocols; Accounting; Risk management and compliance. \\
\addlinespace[4pt]
\textbf{Health \& Wellness} & Medical appointment booking; Biomedical research; Medical information lookup; Fitness tracking; Nutrition and diet; Mental wellness. \\
\addlinespace[4pt]
\textbf{Academic Research} & Earth science and environmental monitoring; Knowledge management; Biomedical research; Literature search; Citation analysis; Data collection and processing; Arithmetic and general calculation; Scientific and symbolic computing; Statistical analysis; Academic writing and collaboration; Data visualization; Experiment design. \\
\addlinespace[4pt]
\textbf{Data Analytics} & Earth science and environmental monitoring; Trend analysis and analytics; NoSQL and document stores; Relational databases; Arithmetic and general calculation; Scientific and symbolic computing; Statistical analysis; Data visualization; Agent orchestration and multi-agent workflows; Model hubs and LLM APIs; Machine learning; Data pipelines; Knowledge graphs and vector retrieval; Web scraping; Report generation. \\
\addlinespace[4pt]
\textbf{Content Creation} & Trend analysis and analytics; Video and audio production; Image generation and editing; Social network integration; Video and content platforms; Writing and copywriting; Graphic design; Content scheduling; Agent orchestration and multi-agent workflows; Model hubs and LLM APIs. \\
\addlinespace[4pt]
\textbf{Software Engineering} & Code development; Package and dependency management; Cloud services and deployment; CI/CD and DevOps; Browser control and interaction; Software testing and quality assurance; NoSQL and document stores; Relational databases; API integration and testing; Shell and infrastructure; Frameworks and libraries; Technical documentation. \\
\addlinespace[4pt]
\textbf{Cybersecurity} & Compliance and audit; Log analysis; Security assessment; Threat intelligence; Network monitoring and response; Identity and access management; Security training. \\
\bottomrule
\end{tabularx}
\caption{Complete environment taxonomy. The table covers all 10 business domains and all 84 unique capability categories. Categories shared across domains are repeated with identical names; the number of domain--capability associations therefore exceeds 84.}
\label{tab:environment-taxonomy}
\end{table*}

\section{Curriculum and Training Details}
\label{app:curriculum-details}

This appendix specifies task partitions, checkpoint initialization, and training objectives for Section~\ref{sec:crossenv-curriculum}. Appendix~\ref{app:analysis-protocol} documents curriculum evaluation, and Appendix~\ref{app:generalization-protocol} documents generalization tests.

\subsection{Curriculum Organization}
\label{app:curriculum-organization}

\noindent\textbf{Task levels.} Tasks are grouped into L1--L5 according to environment count, business-domain scope, and user interaction. L1 contains single-environment tasks, which use the representation in Equation~\ref{eq:crossenv-task} with $|\mathcal E_S|=1$. Within a single business domain, L2 introduces tasks involving two environments, and L3 extends coordination to three or more environments. L4 requires coordination across multiple business domains and three or more environments. L5 retains L4's environment-count and domain-scope requirements and introduces user interaction. Let $\mathcal D$ denote the synthesized training tasks. The task set at each level is
\begin{equation}
\begin{gathered}
\mathcal D_k=\{\tau\in\mathcal D\mid \operatorname{level}(\tau)=L_k\},\\
k=1,\ldots,5.
\end{gathered}
\end{equation}

\subsection{Single-Environment Supervised Fine-Tuning}
\label{app:curriculum-sft}

We perform SFT on reference execution trajectories from $\mathcal D_1$ to teach tool selection, argument specification, and the use of tool outputs to guide subsequent actions~\citep{qin2023toolllmfacilitatinglargelanguage,yang2025toolmindtechnicalreportlargescale,xu2025agenttrekagenttrajectorysynthesis}. For task $\tau$, let $A^*=(a_0^*,\ldots,a_{T_\tau-1}^*)$ denote the reference action sequence, with length $T_\tau=|A^*|$. Here, $I$ denotes the task instruction and $h_t$ contains the preceding interactions. The policy $\pi_\theta$ predicts actions conditioned on $I$ and $h_t$. The training objective is
\begin{equation}
\label{eq:curriculum-sft-objective}
\mathcal L_{\mathrm{SFT}}(\theta)
=-\mathbb E_{\tau\sim\mathcal D_1}\!
\sum_{t=0}^{T_\tau-1}\!\log\!\pi_\theta(a_t^*\mid I,h_t).
\end{equation}
The resulting checkpoint $\theta_1$ initializes subsequent cross-environment RL.

\subsection{Stagewise Reinforcement Learning on Cross-Environment Tasks}
\label{app:curriculum-rl}

\noindent\textbf{Stagewise training.} Starting from the L1 SFT checkpoint, we train sequentially on L2--L5. For $k=2,\ldots,5$, stage $k$ uses $\mathcal D_k$ and is initialized with $\theta_k^{\mathrm{init}}=\theta_{k-1}$. We denote the checkpoint obtained after this stage by $\theta_k$.

\noindent\textbf{Execution feedback.} For a sampled task $\tau$, the agent starts from the joint initial state $\mathbf s_{0,\tau}$ and selects tool calls based on the instruction $I$ and interaction history $h_t$. At L5, execution also includes interaction with a user. Let $A$ denote the agent's action sequence, including tool calls and user-directed interactions where applicable. The verification rules $C_\tau$ evaluate its outputs and final states, producing the execution reward $r_\tau(A)$. The training objective at level $k$ is
\begin{equation}
\label{eq:curriculum-rl-objective}
\begin{gathered}
J_k(\theta)=\mathbb E_{\substack{\tau\sim\mathcal D_k\\A\sim\pi_\theta(\cdot\mid\tau)}}[r_\tau(A)],\\
k=2,\ldots,5.
\end{gathered}
\end{equation}
Here, $\pi_\theta(\cdot\mid\tau)$ denotes the rollout distribution induced by the policy interacting with task $\tau$; the policy observes only the instruction and interaction history, not the underlying environment states or verification rules.

\subsection{Training Configuration}
\label{app:training-configuration}

The mainline uses Qwen3 backbones at 1.7B, 4B, and 8B. Table~\ref{tab:training-data} gives the stagewise data counts. The settings below document L1 SFT and L2--L5 RL: single-environment supervision is followed by two-environment tasks within one domain, three-environment tasks within one domain, three-environment tasks across domains, and tasks requiring user interaction. At each transition, we load the selected policy weights from the preceding stage and reinitialize the optimizer and task sampler. The paper's L4 corresponds to L5 in the historical experiment directories; this naming is distinct from the paper's subsequent user-interaction stage.

\begin{table}[t]
\centering
\fontsize{9.5}{11.5}\selectfont
\setlength{\tabcolsep}{3pt}
\renewcommand{\arraystretch}{1.18}
\begin{tabularx}{\linewidth}{>{\raggedright\arraybackslash}p{0.31\linewidth}>{\centering\arraybackslash}X>{\centering\arraybackslash}X}
\toprule
\rowcolor{black!6}
\textbf{Parameter} & \textbf{L1 SFT} & \textbf{L2--L5 RL} \\
\midrule
Learning rate & $5\times10^{-7}$\newline (peak) & $5\times10^{-7}$\newline (constant) \\
LR schedule & $\approx$10\% warmup,\newline cosine decay & Constant \\
\addlinespace[4pt]
Global batch & 8 assistant\newline responses & 16 task slots \\
Samples per task & Reference\newline supervision & 8 trajectories \\
\addlinespace[4pt]
PPO epochs & --- & 1 \\
KL coefficient & --- & 0.001\newline in the loss \\
Temperature & --- & Train: 1.0\newline Validation: 0 \\
\bottomrule
\end{tabularx}
\caption{Training configuration for L1 SFT and L2--L5 RL. Each RL batch contains 128 sampled trajectories.}
\label{tab:training-hyperparameters}
\end{table}

\noindent\textbf{L5 user interaction.} L5 retains the L4 optimization settings and reward formulation, including the interaction-turn budgets. Qwen3.8-Flash generates user replies during interaction. The agent initially receives only the public task instruction, with user-specific facts withheld. It can request this information by calling \texttt{ask\_user} with a \texttt{question} argument. The reply is appended to the context as a tool-role message, after which the agent continues execution. Each task allows at most three questions. These calls consume assistant turns, context, and wall-clock time, but do not count toward the environment's business-action budget.

\noindent\textbf{Supervised loss.} SFT supervises only the current assistant response. Preceding dialogue and tool feedback are masked from the loss, so the global batch size counts assistant-response samples rather than complete trajectories.

\noindent\textbf{GRPO advantages.} RL uses trajectory-level outcome GRPO. For the eight trajectories sampled for a task, let $R_i$ be trajectory $i$'s terminal reward, and let $\mu_R$ and $\sigma_R$ be the within-group reward mean and standard deviation. The normalized advantage is
\begin{equation}
\label{eq:grpo-trajectory-advantage}
\hat A_i=\frac{R_i-\mu_R}{\sigma_R+10^{-6}}.
\end{equation}
This advantage is assigned to all assistant-generated tokens in that trajectory. The KL term is added to the optimization loss with coefficient 0.001.

\subsection{Terminal Reward}
\label{app:training-reward}

For L2--L5, the terminal reward combines progress on task checks, constraint violations, invalid actions, and completion bonuses:
\begin{equation}
\label{eq:training-terminal-reward}
\begin{aligned}
R={}&0.85(P_T-P_0)-0.20(V_T-V_0)\\
   &-0.01n_{\mathrm{invalid}}+0.10I_F+0.05I_FI_De.
\end{aligned}
\end{equation}
Here, $P$ is the weighted fraction of task checks passed, and $V$ is the weighted fraction of invariant constraints violated; subscripts $0$ and $T$ denote the initial and terminal values. The count $n_{\mathrm{invalid}}$ records invalid actions. The indicator $I_F$ requires all execution checks to pass and a valid final answer, while $I_D$ indicates satisfaction of the tool-discovery protocol. The efficiency term is
\begin{equation}
\label{eq:training-efficiency-reward}
e=\operatorname{clip}\!\left(\frac{T_{\max}-T}{T_{\max}-T_{\mathrm{ref}}},0,1\right),
\end{equation}
where $T$ is the number of interaction turns. The budget parameters $(T_{\max},T_{\mathrm{ref}})$ are $(32,18)$ for L2 and $(40,30)$ for L3--L5. Thus, the reward gives partial credit for progress and is not a binary task-success signal.

\section{Evaluation Model Versions}
\label{app:model-versions}

Table~\ref{tab:model-versions} lists all models in Table~\ref{tab:overall-no-bfcl-mcp}, their sources, and available evaluation identifiers. The GPT-5.6-Luna and Gemini-3.8-Flash API IDs are recorded in the evaluation configurations and request code. IDs marked Claw-Eval are recorded for that benchmark; their use on other benchmarks remains to be verified. API IDs record requested model names; backend snapshots remain unverified. Hugging Face repositories identify public releases, but their evaluation revisions remain to be verified against the run configurations. Agent-World entries include the frozen checkpoint identifiers recorded for our in-house reproductions. For EnvScaler-4B and CrossEnv, we report the Qwen3 backbone; CrossEnv entries also specify the final curriculum stage. The Qwen3.8 link describes a related open-weight release and does not establish equivalence to the Max API.

\begin{table*}[t]
\centering
\fontsize{9}{10}\selectfont
\setlength{\tabcolsep}{6pt}
\renewcommand{\arraystretch}{1.10}
\begin{tabularx}{\linewidth}{>{\raggedright\arraybackslash}p{0.245\linewidth}>{\raggedright\arraybackslash}X}
\toprule
\rowcolor{black!8}
\textbf{Model} & \textbf{Version / source} \\
\midrule
\rowcolor{black!4}
\multicolumn{2}{l}{\textbf{Frontier closed-source models}} \\
GPT-5.6-Luna & \href{https://developers.openai.com/api/docs/models/gpt-5.6-luna}{OpenAI model documentation}\newline API ID: \nolinkurl{gpt-5.6-luna} \\
Claude-Haiku-4.5 & \href{https://www.anthropic.com/news/claude-haiku-4-5}{Anthropic model announcement}\newline Claw-Eval API ID: \nolinkurl{bedrock/claude-haiku-4_5} \\
Gemini-3.8-Flash & \href{https://deepmind.google/models/model-cards/gemini-3-8-flash/}{Google DeepMind model card}\newline API ID: \nolinkurl{gemini-3.8-flash} \\
Qwen3.8-Max & Claw-Eval API ID: \nolinkurl{qwen3.8-max-0902}\newline \href{https://huggingface.co/Qwen/Qwen3.8-2.4T-A95B}{Qwen/Qwen3.8-2.4T-A95B} (related model card) \\
\midrule
\rowcolor{black!4}
\multicolumn{2}{l}{\textbf{Open-weight general-purpose models}} \\
DeepSeek-V4-Pro & \href{https://huggingface.co/deepseek-ai/DeepSeek-V4-Pro-0813}{deepseek-ai/DeepSeek-V4-Pro-0813}\newline \textcolor{black!65}{Recorded ID: \texttt{deepseek-v4-pro-0813}} \\
GLM-5.3 & \href{https://huggingface.co/zai-org/GLM-5.3}{zai-org/GLM-5.3}\newline Claw-Eval API ID: \nolinkurl{glm-5.3} \\
Kimi-K3 & \href{https://huggingface.co/moonshotai/Kimi-K3}{moonshotai/Kimi-K3}\newline Claw-Eval API ID: \nolinkurl{kimi-k3} \\
Qwen3-1.7B & \href{https://huggingface.co/Qwen/Qwen3-1.7B}{Qwen/Qwen3-1.7B} \\
Qwen3-4B & \href{https://huggingface.co/Qwen/Qwen3-4B}{Qwen/Qwen3-4B} \\
Qwen3-8B & \href{https://huggingface.co/Qwen/Qwen3-8B}{Qwen/Qwen3-8B} \\
\midrule
\rowcolor{black!4}
\multicolumn{2}{l}{\textbf{Environment scaling methods}} \\
EnvScaler-1.7B & \href{https://huggingface.co/XXHStudyHard/EnvScaler-Qwen3-1.7B}{XXHStudyHard/EnvScaler-Qwen3-1.7B} \\
EnvScaler-4B & Our reproduction; Qwen3-4B backbone \\
EnvScaler-8B & \href{https://huggingface.co/XXHStudyHard/EnvScaler-Qwen3-8B}{XXHStudyHard/EnvScaler-Qwen3-8B} \\
Arctic-AWM-4B & \href{https://huggingface.co/Snowflake/Arctic-AWM-4B}{Snowflake/Arctic-AWM-4B} \\
Arctic-AWM-8B & \href{https://huggingface.co/Snowflake/Arctic-AWM-8B}{Snowflake/Arctic-AWM-8B} \\
Toucan-7B & \href{https://huggingface.co/Agent-Ark/Toucan-Qwen2.5-7B-Instruct-v0.1}{Agent-Ark/Toucan-Qwen2.5-7B-Instruct-v0.1} \\
EnvACE-1.7B & \href{https://huggingface.co/Team-ACE/EnvACE-Qwen3-1.7B}{Team-ACE/EnvACE-Qwen3-1.7B} \\
EnvACE-8B & \href{https://huggingface.co/Team-ACE/EnvACE-Qwen3-8B}{Team-ACE/EnvACE-Qwen3-8B} \\
Agent-World-4B & Our reproduction; Qwen3-4B RL step50\newline \nolinkurl{AgentWorld-Qwen3-4B-dlcxnsnwrsoefb16-step50} \\
Agent-World-8B & Our reproduction; Qwen3-8B RL step50\newline \nolinkurl{AgentWorld-Qwen3-8B-dlc13s4sujlwtirs-step50} \\
\midrule
\rowcolor{black!4}
\multicolumn{2}{l}{\textbf{CrossEnv}} \\
\textbf{CrossEnv-1.7B} & Trained in this work (Qwen3-1.7B); final L5 checkpoint \\
\textbf{CrossEnv-4B} & Trained in this work (Qwen3-4B); final L5 checkpoint \\
\textbf{CrossEnv-8B} & Trained in this work (Qwen3-8B); final L5 checkpoint \\
\bottomrule
\end{tabularx}

\normalsize
\caption{Model versions and sources.}
\label{tab:model-versions}
\end{table*}

\section{Evaluation Protocol Notes}
\label{app:evaluation-protocol}

\subsection{Benchmark Subsets and Metrics}

\noindent\textbf{$\tau^2$-bench.} We evaluate Retail (114 tasks), Airline (50), Telecom (114), and the Banking Knowledge extension (97)~\citep{shi2026tauknowledge}, for a total of 375 tasks. These domains assess conversational task completion under domain-specific rules; Telecom additionally requires coordination with a user who can act in the environment. We report pass@1.

\noindent\textbf{DeepPlanning-Shopping.} We evaluate the L1, L2, and L3 difficulty subsets, containing 50, 50, and 20 tasks, respectively. These 120 tasks assess information gathering and planning under local and global constraints. We report the average case score. These benchmark difficulty labels are separate from the CrossEnv curriculum levels.

\noindent\textbf{VitaBench.} We evaluate 25 tasks from each of Delivery, In-store, OTA, and the cross-scenario subset, for a total of 100 tasks. They assess service execution, information integration, and adaptation to user requests, with the cross-scenario subset requiring coordination across services. We report pass@1.

\noindent\textbf{Claw-Eval.} We evaluate the adapted text subset of 157 General and 38 Multi-Turn tasks, excluding tasks involving multimodal content. General covers service orchestration, while Multi-Turn covers professional dialogue. We report pass@1; the task-score threshold and protocol adaptations are specified below.

The four benchmarks comprise 790 tasks per model: 375 from $\tau^2$-bench, 120 from DeepPlanning, 100 from VitaBench, and 195 from Claw-Eval. Within each benchmark, Avg. is weighted by subset size. Capability groups describe the main evaluation emphasis rather than mutually exclusive abilities: VitaBench and Claw-Eval Multi-Turn also involve user interaction.

\subsection{Evaluation Settings}

\noindent\textbf{Checkpoints and decoding.} We evaluate CrossEnv models after L1--L5 training. Capability tracking extends through L5. The Mixed comparison and longer-workflow tests use checkpoints through L4. Agent-World-4B and Agent-World-8B use our reproduced Qwen3 RL step50 checkpoints. The recorded October baseline campaigns use temperature 0, thinking enabled, and no replay of historical reasoning; Toucan uses its native Qwen2.5 non-thinking template. GPT-5.6-Luna and Claude-Haiku-4.5 use temperature 1 on $\tau^2$-bench, DeepPlanning, and VitaBench. Claw-Eval settings are specified separately below.

\noindent\textbf{$\tau^2$-bench.} The four API reference models Qwen3.8-Max, DeepSeek-V4-Pro, GLM-5.3, and Kimi-K3 use a 128K context budget. Qwen3 Base, EnvScaler, Arctic-AWM, and Toucan use 64K. Local runs use the Hermes tool-call adapter with explicit model-format handling. Fixed-budget failures receive zero credit and remain in the task denominator.

\noindent\textbf{VitaBench.} October runs use a 4,096-token judge output budget. The EnvACE-1.7B and Arctic-AWM-4B runs use a 16,384-token judge budget and a YaRN original-position baseline of 40,960. Tool outputs and user interactions are scored using the benchmark's task rubrics.

\noindent\textbf{Claw-Eval scoring.} We perform one trial per task on the 157 General and 38 Multi-Turn tasks specified above. A native task score of at least 0.75 counts as success. All 195 tasks remain in the denominator, including model errors and fixed-budget failures. The native score is
\begin{equation}
\begin{aligned}
q&=\mathrm{safety}\,(0.8\,\mathrm{completion}\\
 &\hspace{4.4em}+0.2\,\mathrm{robustness}),\\
\mathrm{task\_score}&=\operatorname{round}(q,4).
\end{aligned}
\end{equation}
Gen. and MT report subset success rates, and the table's average is
\begin{equation}
\mathrm{Avg.}=\frac{157s_{\mathrm{Gen.}}+38s_{\mathrm{MT}}}{195}.
\end{equation}
The mean native task score is a separate metric from this success rate.

\noindent\textbf{Claw-Eval local models.} Qwen3 and EnvScaler use temperature 0, thinking enabled, no replay of historical reasoning, a 131,072-token context, and an 8,192-token Agent output budget. Toucan-7B, EnvACE-1.7B/\allowbreak{}8B, and Arctic-AWM-4B/\allowbreak{}8B use the same context and Agent output budgets; Toucan uses its native non-thinking template. Native per-task turn and time limits are retained. The judge and user simulator use Qwen3.8-Max with output budgets of 8,192 and 65,536 tokens, respectively. Toucan-7B, EnvACE-8B, and Arctic-AWM-8B use eight task workers; Arctic-AWM-4B uses two TP1 replicas with four workers each.

\noindent\textbf{Claw-Eval API models.} The recorded identifiers for Qwen3.8-Max, DeepSeek-V4-Pro, GLM-5.3, and Kimi-K3 are \texttt{qwen3.8-max-0902}, \texttt{deepseek-v4-pro-0813}, \texttt{glm-5.3}, and \texttt{kimi-k3}. These models use temperature 0, thinking enabled, and no replay of historical reasoning. Agent output limits follow upstream defaults, with a 131,072-token input-plus-output budget per response; judge and user output budgets are 8,192 and 65,536 tokens. Full evaluations use eight workers per model. GPT-5.6-Luna uses temperature 1 and Responses medium reasoning. Gemini-3.8-Flash uses temperature 0 with native signature handling. Claude-Haiku-4.5 uses the \texttt{bedrock/claude-haiku-4\_5} Chat-compatible API at temperature 0 with thinking requested. These three models have no explicit Agent output cap and use the same 131,072-token input-plus-output budget, with at most eight task workers per model.

\noindent\textbf{Claw-Eval tool interfaces.} Tasks run in separate gVisor environments. IQS replaces Google via Novada search. OCR and long-document text are retained in files and exposed through previews and segmented reads; the Read preview limit is 16,000 characters. Adapter configurations include model-specific tool-call parsing, web-fetch transport handling, PDF rendering, and non-UTF-8 output handling. The revised web-fetch configuration uses a cancellable total deadline of at most 30 seconds and a 35-second outer RPC wait.

\section{Curriculum Learning Analysis}
\label{app:analysis-protocol}

We define the metrics used in Section~\ref{sec:curriculum-analysis}. Each metric uses a fixed evaluation set across checkpoints. For a set of evaluated requirements $\mathcal U$, let $b_m(u)\in\{0,1\}$ indicate whether model $m$ satisfies requirement $u$, and define
\begin{equation}
\label{eq:capability-requirement-rate}
r_m(\mathcal U)=\frac{100}{|\mathcal U|}\sum_{u\in\mathcal U}b_m(u).
\end{equation}

\subsection{Stagewise Capability Metrics}

\noindent\textbf{Tool execution coverage.} This measures whether the agent executes the tools required by reference workflows, capturing basic tool-use coverage. For the evaluated reference tool requirements $\mathcal U_{\mathrm{tool}}$,
\begin{equation}
S_{\mathrm{tool}}(m)=r_m(\mathcal U_{\mathrm{tool}}).
\end{equation}
Coverage does not by itself establish correct arguments or task success.

\noindent\textbf{Complete environment coverage.} This measures whether the agent acts in every required environment, detecting tasks where it omits an entire service. For task $\tau$, let $\mathcal E_\tau$ and $\widehat{\mathcal E}_\tau(m)$ be the required and actually used environment sets. On evaluation tasks $\mathcal D_{\mathrm{env}}$,
\begin{equation}
S_{\mathrm{env}}(m)=\frac{100}{|\mathcal D_{\mathrm{env}}|}
\sum_{\tau\in\mathcal D_{\mathrm{env}}}
\mathbf 1[\mathcal E_\tau\subseteq\widehat{\mathcal E}_\tau(m)].
\end{equation}

\noindent\textbf{Late-workflow tool coverage.} This measures execution of tool requirements in the later portion of reference workflows, assessing whether the agent continues beyond early steps. For the evaluated late-workflow requirements $\mathcal U_{\mathrm{late}}$,
\begin{equation}
S_{\mathrm{late}}(m)=r_m(\mathcal U_{\mathrm{late}}).
\end{equation}

\noindent\textbf{Target-state check pass rate.} This measures whether execution produces the required business states, complementing action coverage with outcome correctness. For target-state checks $\mathcal U_{\mathrm{state}}$ evaluable across the compared models,
\begin{equation}
S_{\mathrm{state}}(m)=r_m(\mathcal U_{\mathrm{state}}).
\end{equation}
Unavailable checks are excluded consistently across models, rather than counted as failures.

\subsection{User Interaction Metric}
\label{app:user-interaction-metric}

This metric assesses three aspects of user interaction: obtaining required information through a relevant question, using the reply in appropriate tool arguments with normal tool execution, and passing native strict task checks. Their combination captures intermediate progress as well as task completion. Let $n_q$, $n_u$, and $n_s$ count tasks satisfying these criteria on the same $N=40$ held-out interaction tasks:
\begin{equation}
\label{eq:user-interaction-score}
S_{\mathrm{int}}=100\frac{n_q+n_u+n_s}{N}.
\end{equation}
The score ranges from 0 to 300. The criteria may hold on different tasks, so it is neither a success percentage nor the rate of completing all three steps together.

\subsection{Curriculum and Mixed RL Metrics}

Both schedules initialize from the same Qwen3-8B L1 SFT checkpoint. The compared curriculum and Mixed checkpoints use 243.04 and 238.85 cumulative RL GPU-hours, respectively. Mixed training samples from the L2, L3, and L4 pools of 579, 620, and 600 tasks in a $1{:}1{:}1$ ratio. Each batch contains eight tasks from each level, for 24 tasks in total, with eight rollouts per task (192 rollouts per batch). Tasks are shuffled independently within each level, and the combined batch is shuffled again. The runs differ in parallelism, batch configuration, and reference-model settings; this comparison therefore assesses the two training schedules at similar compute budgets under their respective configurations, rather than isolating curriculum order.

Strict success measures complete task attainment; cumulative rollouts measure sampling volume at a given compute budget. Let $y_m(\tau)$ indicate that all task requirements pass and $n_j$ be the rollout count for completed update $j$. We report
\begin{equation}
\begin{aligned}
\mathrm{Success}(m)&=\frac{100}{|\mathcal D|}\sum_{\tau\in\mathcal D}y_m(\tau),\\
\mathrm{Rollouts}(h)&=\sum_{j:\,h_j\le h}n_j,
\end{aligned}
\end{equation}
where $h_j$ is cumulative RL GPU-hours at update $j$. The comparison uses the same 180 validation tasks at similar GPU-hours. Rollout volume does not measure trajectory validity or task success; the comparison reflects the current training configurations.

\section{Generalization Analysis}
\label{app:generalization-protocol}

We define the tests and metrics used in Section~\ref{sec:generalization}. Combination tests evaluate transfer to new environment pairs, and long-horizon tests evaluate transfer to longer workflows; task-similarity analysis compares CrossEnv and Base at 4B and 8B.

\subsection{Combination Generalization}

This test asks whether an agent can coordinate familiar environments in new combinations. Seen pairs co-occur in training; unseen pairs never co-occur, including within larger training combinations. Each group contains 150 tasks from 30 environment pairs. Strict success uses all tasks, retaining model failures in the denominator. Let $y_m(\tau)$ indicate that model $m$ passes all task requirements. For group $g$,
\begin{equation}
S_g(m)=\frac{100}{|\mathcal D_g|}\sum_{\tau\in\mathcal D_g}y_m(\tau).
\end{equation}
Table~\ref{tab:generalization-combinations} reports $S_g(\mathrm{CrossEnv})-S_g(\mathrm{Base})$ in percentage points.

Observation, State, and Answer scores localize gains to business observations, final states, and final answers. For category $c$, let $\mathcal D_{g,c}$ contain applicable tasks with native scores available for both models, and $\mathcal C_{\tau,c}$ contain the task's category-specific hard checks. With binary check outcomes $b_m(\tau,k)$,
\begin{equation}
S_{g,c}(m)=\frac{100}{|\mathcal D_{g,c}|}
\sum_{\tau\in\mathcal D_{g,c}}\prod_{k\in\mathcal C_{\tau,c}}b_m(\tau,k).
\end{equation}
Category gains are also CrossEnv minus Base. They require all checks in a category to pass and describe the jointly scored subset, not the full test population.

\subsection{Long-Horizon Generalization}

This test measures transfer to workflows longer than those used in training. Two sets of 40 tasks have reference lengths of 16--18 and 24--28 tool calls, exceeding the training maximum of 14. To capture partial progress, we average each task's fraction of passed observation and target-state checks:
\begin{equation}
\mathrm{Score}(m)=\frac{100}{|\mathcal D|}\sum_{\tau\in\mathcal D}
\frac{\sum_{k\in\mathcal C_\tau}b_m(\tau,k)}{|\mathcal C_\tau|},
\end{equation}
where $\mathcal C_\tau$ contains those checks. Tasks receive equal weight; this is not strict task success. Reference length describes the supplied solution, not a minimum for all valid solutions.

\subsection{External-Task Similarity}

Following the similarity-analysis motivation of EnvScaler~\citep{song2026envscalerscalingtoolinteractiveenvironments}, we test whether gains are concentrated on instructions close to training data. We encode instructions with \nolinkurl{sentence-transformers/all-MiniLM-L6-v2}. For test instruction $x$, training-instruction set $\mathcal X_{\mathrm{train}}$, and embedding function $f$, similarity is
\begin{equation}
s(x)=\max_{z\in\mathcal X_{\mathrm{train}}}
\frac{f(x)^\top f(z)}{\|f(x)\|\,\|f(z)\|}.
\end{equation}
For normalized native task scores $q_m(x)$, per-task and similarity-group gains are
\begin{equation}
\begin{aligned}
\Delta(x)&=100\bigl(q_{\mathrm{CrossEnv}}(x)-q_{\mathrm{Base}}(x)\bigr),\\
\overline\Delta_b&=\frac{1}{|\mathcal B_b|}\sum_{x\in\mathcal B_b}\Delta(x),
\end{aligned}
\end{equation}
where $\mathcal B_b$ is one of four similarity-quantile groups. Figure~\ref{fig:vita-similarity} shows group means with 95\% confidence intervals. The VitaBench-8B panel uses 71 paired tasks, native rubric pass fractions, and paired-task bootstrap intervals; this score differs from strict task success. Results describe the plotted subsets, and instruction similarity alone does not establish unseen environments or domains.

\section{Cross-Environment Training Analysis}
\label{app:crossenv-value}

This appendix defines the execution diagnostics in Table~\ref{tab:crossenv-training-performance}. SingleEnv and CrossEnv RL start from a shared SFT checkpoint and differ in whether their training tasks involve single or multiple environments. Base execution scores are historical references.

\subsection{Execution Diagnostics and Measurement Rules}

The three diagnostics describe data dependencies, continuation of execution, and final outcomes. They are related views: one downstream failure can affect multiple metrics, so they do not identify three independent mechanisms.

\noindent\textbf{Use of prior tool outputs.} Annotate required cross-service dependencies from task instructions and verification rules, then check whether actual objects, identifiers, amounts, or states returned by earlier services are correctly bound to later operation parameters. Both models must use the same applicable tasks and dependency edges. Failing to obtain required information or omitting its downstream operation counts as unsatisfied. Any conditional rate given successful upstream retrieval must identify that restricted denominator.

\noindent\textbf{Later-step completion.} Pre-annotate the business operations required after earlier steps, including retries, notifications, and actions after clearing a blockage. An operation counts as complete only if it executes successfully and produces the required state change. A tool call, mention of its name, or verbal completion claim alone is insufficient; omitted necessary operations remain unsatisfied.

\noindent\textbf{All task constraints met.} On tasks with joint goals, require all necessary service outcomes and constraints to hold simultaneously, including correct orders and notifications, specified cleanup objects, and restrictions on extra writes. Score each task by the conjunction of its requirements rather than the mean fraction of atomic checks passed.

Applicable tasks, dependencies, actions, and constraints use the same annotation rules and denominators across models. Infrastructure anomalies and missing scores are recorded separately; execution failures and omitted required operations remain in the denominator.

\section{Synthesis and Verification Quality}
\label{app:synthesis-quality}

This appendix supplements Section~\ref{sec:synthesis-quality-cost} with the current task-quality evaluation and construction-cost accounting.

\subsection{Execution and Verification Checks}

We sample 100 retained L5 tasks, with 50 each from MCP and Skill environments and stratification by domain. All reference workflows execute successfully and receive full scores. On the tested subsets, replay from a reset environment reproduces these outcomes, omission of a required operation prevents a full score, and semantically equivalent query reorderings preserve full scores. These checks concern reference execution, required operations, and equivalent query orders.

\subsection{Construction-Cost Accounting}

Figure~\ref{fig:task-yield-build-cost} compares candidate output yield as the environment pool grows and average environment-construction cost per output. CrossEnv composes new environments with existing ones and reuses their data, tools, and entity relations. The reported CrossEnv pool of 100 environments yields 51,566 candidate tasks, with a recorded environment-construction cost of 206,496,276 tokens. The average cost is therefore
\begin{equation}
\label{eq:construction-cost-per-candidate}
\frac{206{,}496{,}276}{51{,}566}\approx 4{,}004.5
\end{equation}
tokens per candidate, consistent with the approximately 4.0k value in Section~\ref{sec:synthesis-quality-cost}.

CrossEnv and AgentWorld outputs are screened candidates; EnvScaler outputs are released scenarios. AgentWorld values come from our reproduction, whereas EnvScaler values are estimates derived from its reported figures. Costs cover environment construction rather than end-to-end task generation, and exclude subsequent task instantiation and acceptance. The denominator is candidate output, not the final set of training examples. Dashed extensions in the figure indicate projections under unchanged mean yield and construction cost per additional environment.

\section{Environment and Task Synthesis Comparison Criteria}
\label{app:environment-comparison-criteria}

Table~\ref{tab:environment_comparison} compares explicitly reported synthesis capabilities rather than empirical guarantees or only the default training recipe. A supported optional branch counts as supported; an unsupported mark means that the described pipeline does not explicitly provide the capability.

\noindent\textbf{Executable (Exec.).} Actions run through real services or executable code. Execution during data collection counts, without implying that an online RL training environment is released.

\noindent\textbf{Verifiable (Verif.).} Executable checks evaluate task outcomes, rather than only syntax or tool functionality. Pure-code verification branches count even when another branch uses an LLM judge.

\noindent\textbf{Controllable (Ctrl.).} Explicit settings adjust task complexity through tool count, graph length, branching, constraints, environment count, or discrete complexity levels. Limited support denotes qualitative workflow expansion or adaptive difficulty targeting; a fixed request for diverse or challenging tasks does not constitute an adjustable control.

\noindent\textbf{Cross-environment (Cross).} A synthesized task coordinates multiple environments, applications, or services. AWM's limited support covers bundled service suites, without demonstrated general composition across independently constructed environments.

\noindent\textbf{Consistent (Consist.).} The pipeline explicitly aligns entities or data across environments, or enforces cross-environment constraints. Local database consistency alone does not establish this capability.

\noindent\textbf{Extensible (Ext.).} New environments or services can be integrated with reusable existing ones to generate joint tasks. Interface-based integration counts; integration need not require zero adaptation.

\end{document}